\documentclass[10pt,twocolumn,letterpaper]{article}

\usepackage{iccv}              % To produce the CAMERA-READY version
\usepackage{kotex}
\usepackage{bm} % To produce the REVIEW version
\usepackage[export]{adjustbox}
\usepackage{multirow}
\usepackage{subcaption}
\usepackage{graphicx}
\newcommand{\Fref}[1]{Figure.~\ref{#1}}

\usepackage{booktabs} 
\newcommand{\figteaser}{
  \begin{figure}[t]
    \centering
    \includegraphics[width=0.98\linewidth]{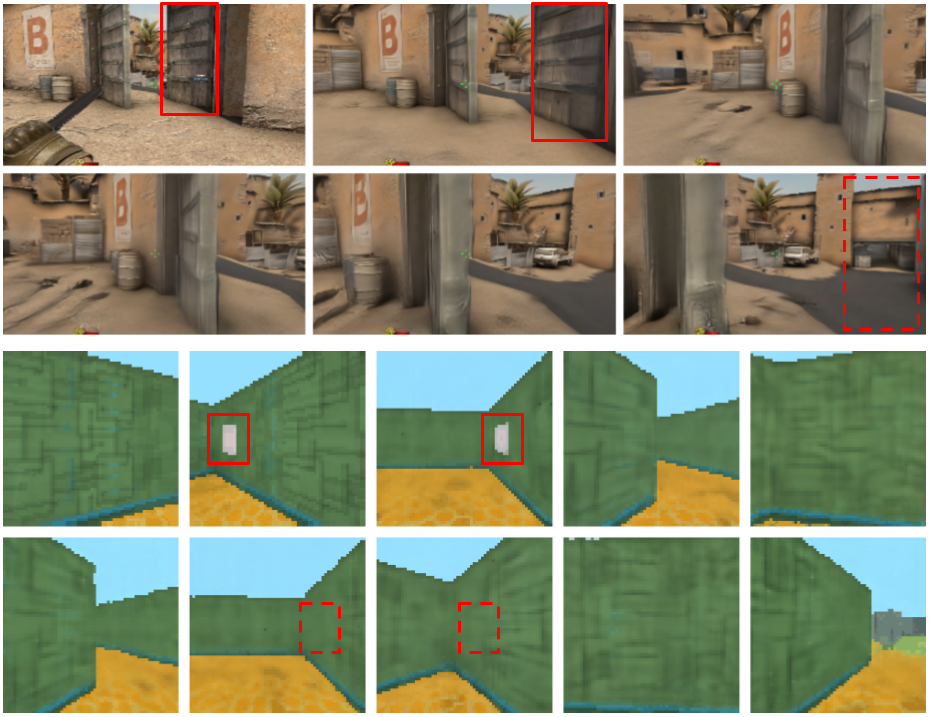}
    \caption{\textbf{Revisiting the same place by a world model.} When prompting the model to revisit the previously synthesized location, the newly generated environment fails to preserve key objects within the red box, such as a door or a picture frame. This inconsistency can be critically detrimental to both agent performance and user experience.}
    \vspace{-6mm}
    \label{fig:main_figure}
  \end{figure}%
}

\newcommand{\figactanalysis}{
  \begin{figure}[t]
    \centering    \includegraphics[width=0.6\linewidth]
    {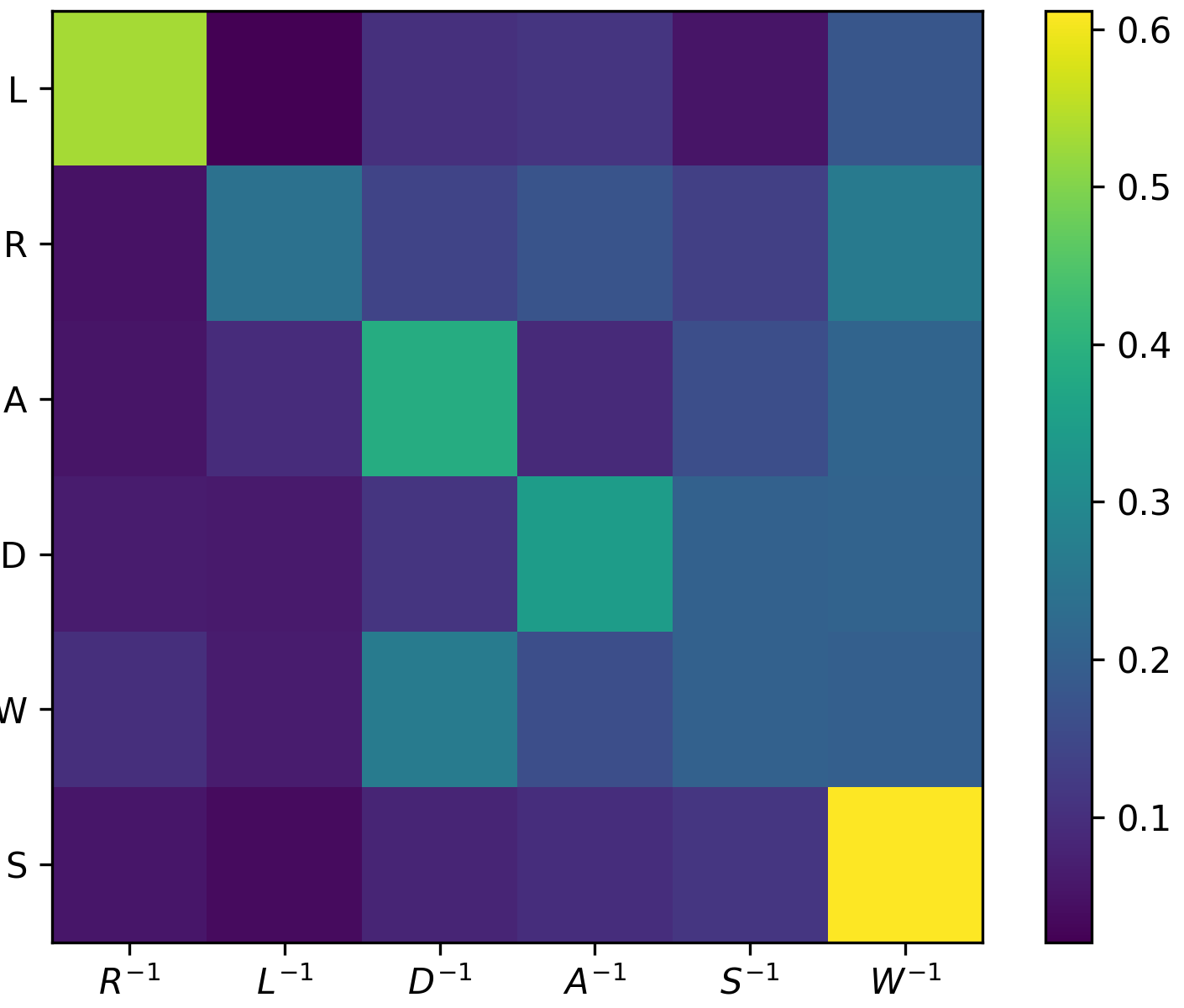}
    \vspace{-1mm}
    \caption{\textbf{Similarity matrix of action and inverse action embeddings} learned from the CS:GO dataset by \textit{IRP}. High diagonal similarity indicates successful inverse relationship learning.}
    \vspace{-6mm}
\label{fig:act_analysis}
  \end{figure}%
}

\newcommand{\figseqlen}{
  \begin{figure}[ht]
    \centering
    \includegraphics[width=0.85\linewidth]
    {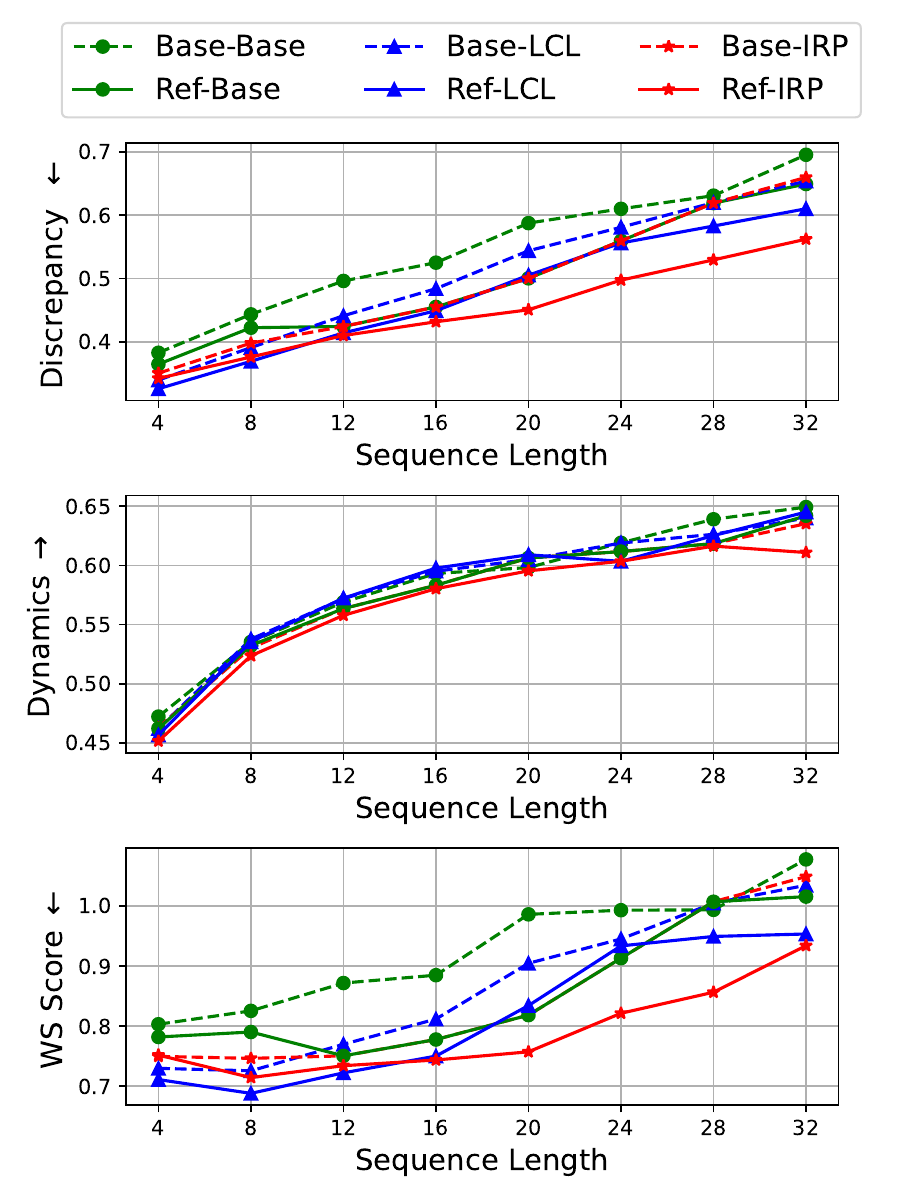}
    \vspace{-2mm}
    \caption{\textbf{Ablation study on sequence length.} We investigate how the sequence length $|\mathcal{A}|$ affects the two components of the World Stability (WS) score: discrepancy and dynamics. The experiments are conducted on the CS:GO dataset. The line plots depict the trends of discrepancy, dynamics, and following WS score across different sequence lengths. 
    }
    \vspace{-6mm}
    \label{fig:abl_seqlen}
  \end{figure}%
}

\newcommand{\figsampling}{
  \begin{figure}[t]
    \centering
    \includegraphics[width=0.8\linewidth]
    {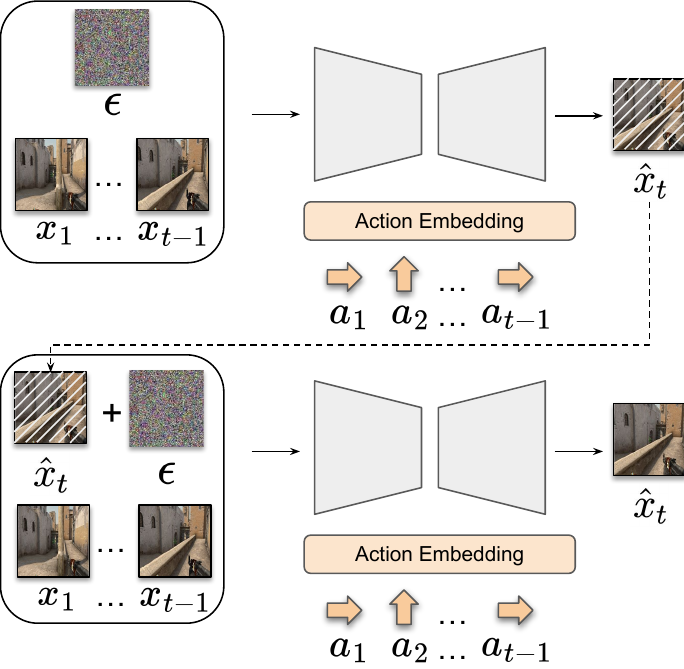}
    \caption{\textbf{Graphical description of Refinement Sampling.} We first sample an observation and inject additional noise into the generated one. Then, the model accepts the noised input and the same conditions for synthesizing a new observation.} 
    \vspace{-6mm}
    \label{fig:sampling}
  \end{figure}%
}

\newcommand{\figexample}{
    \begin{figure*}[h]
    \centering
    \setlength{\tabcolsep}{1pt}
        \begin{tabular}{c c c c c c c c}
        \raisebox{0.01\height}{\rotatebox{90}{\shortstack{\scriptsize Base-Base \\ \scriptsize (WS=1.07)}}} &
        \adjincludegraphics[clip,width=0.13\linewidth,trim={0 0 0 0}]{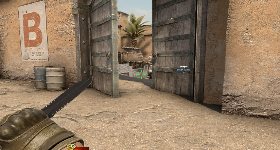} &
        \adjincludegraphics[clip,width=0.13\linewidth,trim={0 0 0 0}]{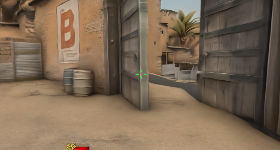} &
        \adjincludegraphics[clip,width=0.13\linewidth,trim={0 0 0 0}]{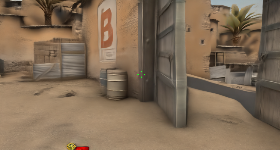} &
        \adjincludegraphics[clip,width=0.13\linewidth,trim={0 0 0 0}]{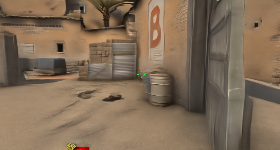} &
        \adjincludegraphics[clip,width=0.13\linewidth,trim={0 0 0 0}]{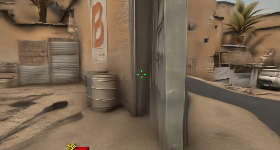}  &

        \adjincludegraphics[clip,width=0.13\linewidth,trim={0 0 0 0}]{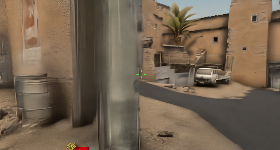}  &
        \adjincludegraphics[clip,width=0.13\linewidth,trim={0 0 0 0}]{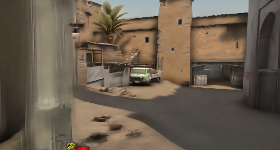} \\ 
        \raisebox{0.01\height}{\rotatebox{90}{\shortstack{\scriptsize Base-LCL \\ \scriptsize (WS=1.09)}}} &
        \adjincludegraphics[clip,width=0.13\linewidth,trim={0 0 0 0}]{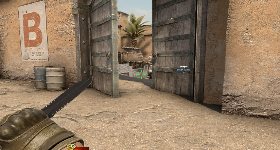} &
        \adjincludegraphics[clip,width=0.13\linewidth,trim={0 0 0 0}]{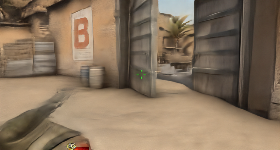} &
        \adjincludegraphics[clip,width=0.13\linewidth,trim={0 0 0 0}]{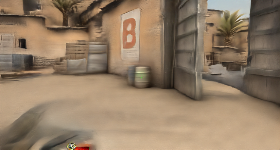} &
        \adjincludegraphics[clip,width=0.13\linewidth,trim={0 0 0 0}]{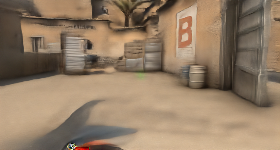} &
        \adjincludegraphics[clip,width=0.13\linewidth,trim={0 0 0 0}]{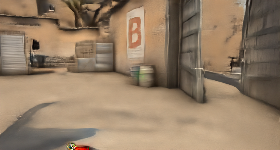}  &

        \adjincludegraphics[clip,width=0.13\linewidth,trim={0 0 0 0}]{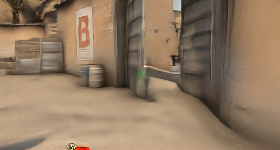}  &
        \adjincludegraphics[clip,width=0.13\linewidth,trim={0 0 0 0}]{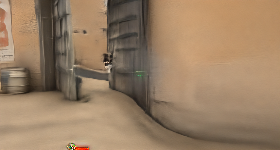} \\ 
        \raisebox{0.01\height}{\rotatebox{90}{\shortstack{\scriptsize Base-IRP \\ \scriptsize (WS=1.03)}}} &
        \adjincludegraphics[clip,width=0.13\linewidth,trim={0 0 0 0}]{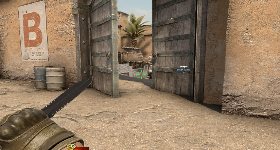} &
        \adjincludegraphics[clip,width=0.13\linewidth,trim={0 0 0 0}]{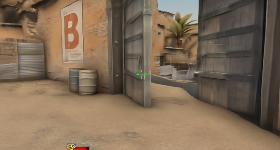} &
        \adjincludegraphics[clip,width=0.13\linewidth,trim={0 0 0 0}]{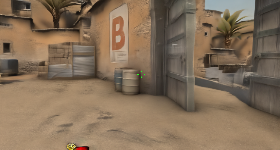} &
        \adjincludegraphics[clip,width=0.13\linewidth,trim={0 0 0 0}]{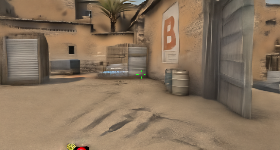} &
        \adjincludegraphics[clip,width=0.13\linewidth,trim={0 0 0 0}]{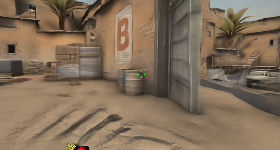}  &

        \adjincludegraphics[clip,width=0.13\linewidth,trim={0 0 0 0}]{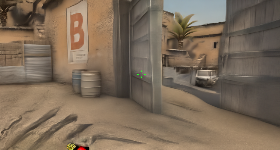}  &
        \adjincludegraphics[clip,width=0.13\linewidth,trim={0 0 0 0}]{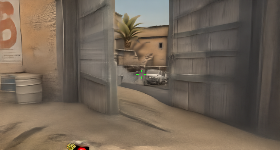} \\
        \raisebox{0.01\height}{\rotatebox{90}{\shortstack{\scriptsize Ref-IRP \\ \scriptsize (WS=0.99)}}} &
        \adjincludegraphics[clip,width=0.13\linewidth,trim={0 0 0 0}]{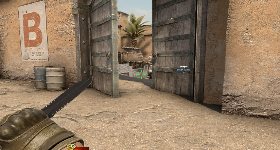} &
        \adjincludegraphics[clip,width=0.13\linewidth,trim={0 0 0 0}]{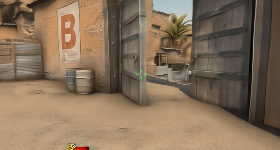} &
        \adjincludegraphics[clip,width=0.13\linewidth,trim={0 0 0 0}]{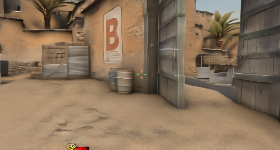} &
        \adjincludegraphics[clip,width=0.13\linewidth,trim={0 0 0 0}]{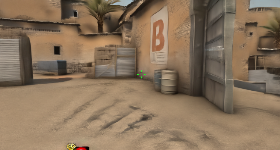} &
        \adjincludegraphics[clip,width=0.13\linewidth,trim={0 0 0 0}]{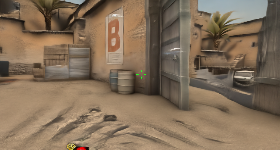}  &

        \adjincludegraphics[clip,width=0.13\linewidth,trim={0 0 0 0}]{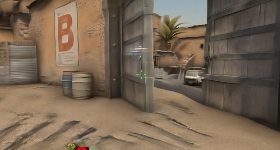}  &
        \adjincludegraphics[clip,width=0.13\linewidth,trim={0 0 0 0}]{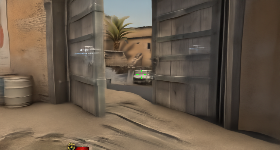} \\ 
 
        \end{tabular}
    \vspace{-2mm}
    \caption{\textbf{Qualitative evaluation with the World Stability score on CS:GO.} To evaluate world stability, we follow the proposed evaluation protocol by rotating the camera position left by a certain angle and then rotating it back to the original position by the same amount.  A smaller discrepancy between the first and last frames indicates higher world stability.
    In the Base model(\textit{Base-Base}), unintended position shifts occur, and the door disappears. \textit{Base-LCL} preserves the door but introduces significant distortions. \textit{Base-IRP} produces the most stable results, while refinement sampling(\textit{Ref-IRP}) further aligns the door's position with the ground truth and improves details such as texture quality. Notably, these qualitative observations align with the WS scores, demonstrating that WS score effectively measures world stability.
    }
    \vspace{-4mm}
    \label{fig:qual_csgo}
\end{figure*}
}

\newcommand{\figmethodnew}{
    \begin{figure*}[t]
      \begin{minipage}[t]{0.475\linewidth}
        \centering
        \includegraphics[width=\textwidth]{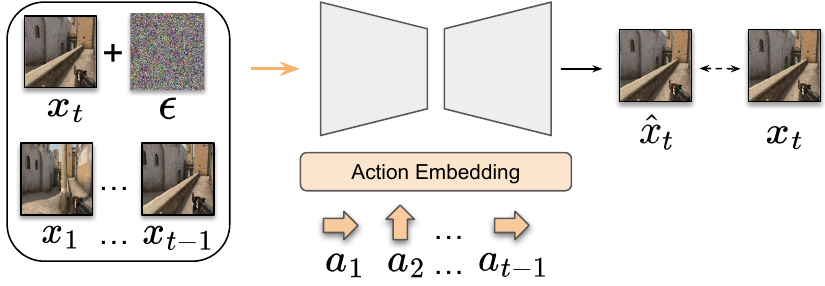}
        \subcaption{Baseline}\label{fig:methoda}
      \end{minipage}
      \hfill
      \begin{minipage}[t]{0.475\linewidth}
        \centering
        \includegraphics[width=\textwidth]{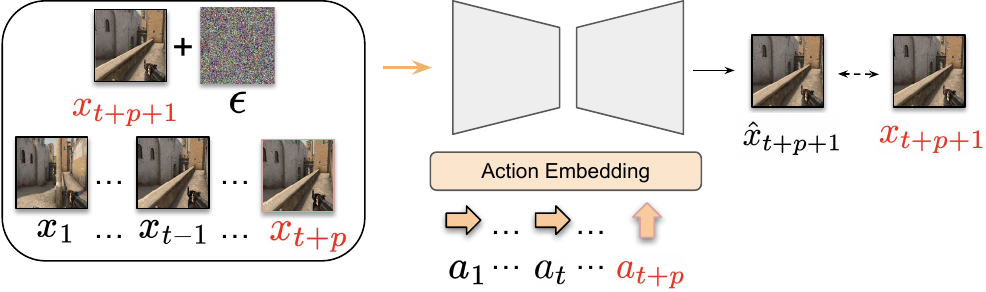}
        \subcaption{Longer Context Length}\label{fig:methodb}
      \end{minipage}
      \\
      \begin{minipage}[t]{0.475\linewidth}
        \centering
        \includegraphics[width=\textwidth]{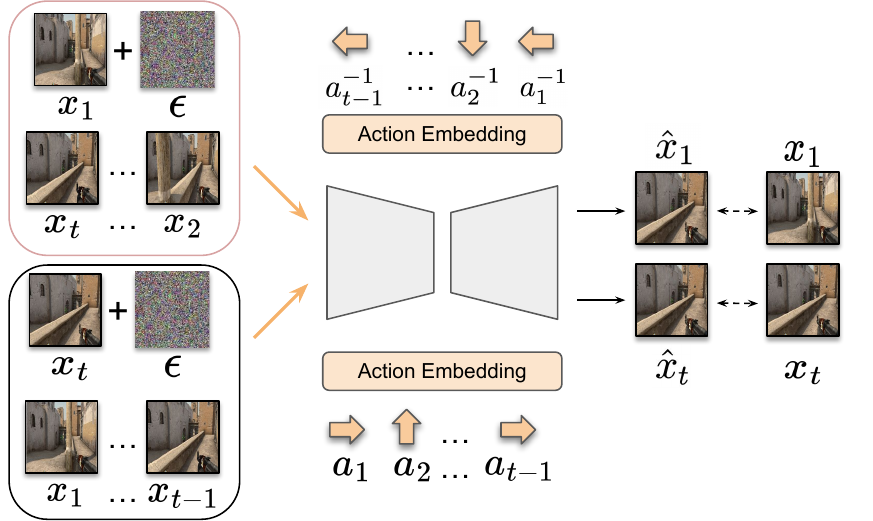}
        \subcaption{Data Augmentation}\label{fig:methodc}
      \end{minipage}
      \hfill
      \begin{minipage}[t]{0.475\linewidth}
        \centering
        \includegraphics[width=\textwidth]{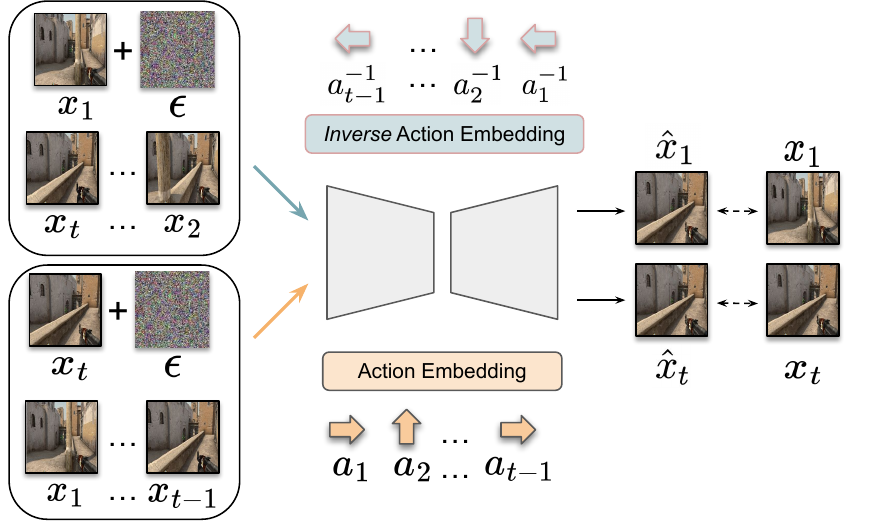}
        \subcaption{Inject Reverse Prediction}\label{fig:methodd}
      \end{minipage}
    \vspace{-2mm}
    \caption{\textbf{Overview of the possible solutions for improving world stability.} The orange and blue arrows indicate that the input uses action embeddings of the same type (or color) to represent the action condition. Red boxes denote the newly introduced components.}
    \label{fig:methodnew}
    \vspace{-4mm}
    \end{figure*}
}

\newcommand{\figdmlabws}{
\begin{figure*}
    \centering
    \begin{subfigure}[b]{0.33\textwidth}
         \centering
         \includegraphics[width=\textwidth]{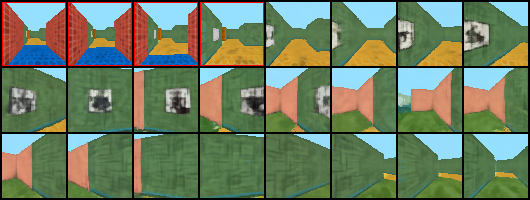}
         \caption{Base-Base (WS=1.58)}
    \end{subfigure}
    \begin{subfigure}[b]{0.33\textwidth}
         \centering
         \includegraphics[width=\textwidth]{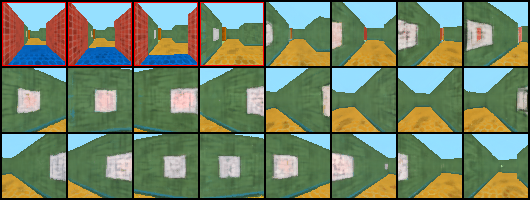}
         \caption{Base-DA (WS=1.40)}
    \end{subfigure}
    \begin{subfigure}[b]{0.33\textwidth}
         \centering
         \includegraphics[width=\textwidth]{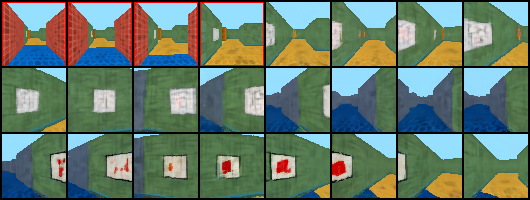}
         \caption{Base-IRP (WS=0.85)}
    \end{subfigure}
    \begin{subfigure}[b]{0.33\textwidth}
         \centering
         \includegraphics[width=\textwidth]{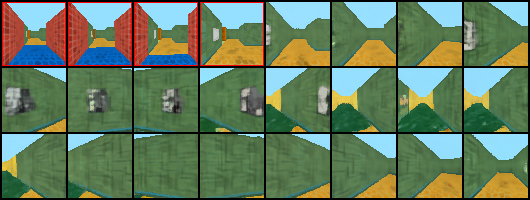}
         \caption{Ref-Base (WS=0.91)}
    \end{subfigure}
    \begin{subfigure}[b]{0.33\textwidth}
         \centering
         \includegraphics[width=\textwidth]{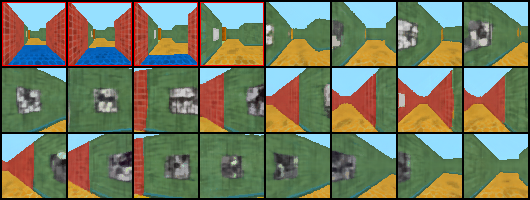}
         \caption{Ref-DA (WS=1.13)}
    \end{subfigure}
    \begin{subfigure}[b]{0.33\textwidth}
         \centering
         \includegraphics[width=\textwidth]{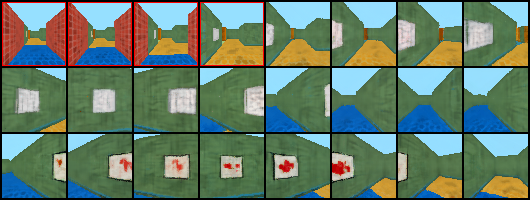}
         \caption{Ref-IRP(WS=0.81)}
    \end{subfigure}
    \caption{\textbf{Qualitative evaluation with the World Stability score on DMLab.} To evaluate world stability, after the given four actions corresponding to the first four frames with red edge, we follow the proposed evaluation protocol by rotating the camera position left by a certain angle and then rotating it back to the original position by the same amount. With the Base method (Base-Base and Ref-Base), the picture frame on the wall completely disappears. However, the Ref-Base model preserves the floor's color better than the Base-Base model. In contrast, other models effectively maintain both the picture frame on the wall and the floor's color.
    } 
    \label{fig:dmlabws}
    \vspace{-4mm}
\end{figure*}
}

\newcommand{\tabmain}{
\begin{table*}[t]
    \centering
    \caption{\textbf{Comparison of World Stability (WS) Score for world generative models using the proposed evaluation framework}.
    We employ three metrics-LPIPS, MEt3R, and DINO features-to measure semantic distance as the basis for WS score.
    Additionally, we report commonly used generative model evaluation metrics, including MSE, PSNR, and SSIM, to assess overall generation quality.
    ``LCL'', ``IRP'', and ``DA'' denote long context length, injected reverse prediction, and data augmentation, respectively. 
    \textbf{Bold} values indicate the best performance across all settings, while \underline{underlined} values highlight the best performance within the base sampling category.
    To the best of our knowledge, this is the first study to evaluate the world stability of state-of-the-art world generative models."
    }
    \vspace{-2mm}
    \resizebox{\textwidth}{!}{
    \begin{tabular}{l|l|l|ccccccc}
    \hline
    \multicolumn{1}{c|}{\multirow{2}{*}{Environment}} & \multicolumn{1}{c|}{\multirow{2}{*}{Sampling}} & \multicolumn{1}{c|}{\multirow{2}{*}{Method}} & \multicolumn{7}{c}{Metrics}                                                                                                                                                            \\ \cline{4-10} 
    \multicolumn{1}{c|}{}                             & \multicolumn{1}{c|}{}                          & \multicolumn{1}{c|}{}                        & \multicolumn{1}{c|}{WS-LPIPS$\downarrow$} & \multicolumn{1}{c|}{WS-MEt3R$\downarrow$} & \multicolumn{1}{c|}{WS-DINO$\downarrow$} & \multicolumn{1}{c|}{FVD$\downarrow$} & \multicolumn{1}{c|}{MSE$\downarrow$} & \multicolumn{1}{c|}{PSNR$\uparrow$} & SSIM$\uparrow$ \\ \hline\hline %\midrule\midrule
    \multirow{6}{*}{CS:GO}                            & \multirow{3}{*}{Base}                          & (\textit{a}) Base                                         &          0.8791                     &              0.7618                  & \multicolumn{1}{c|}{0.8702}        &     611.5                     &       0.1678                   &      13.9723                     & 0.1971     \\
                                                      &                                                & (\textit{b}) LCL                                          &       0.8159                        &    \underline{0.7460}                           & \multicolumn{1}{c|}{\underline{0.8485}}        &        \underline{609.7}                  &          0.1461                &         14.5170                &  \underline{0.2597}   \\
                                                      &                                                & (\textit{c}) IRP                                          &    \underline{0.7774}                           &             0.7583                   & \multicolumn{1}{c|}{0.8608}        &           610.5               &       \underline{0.1431}                   &         \underline{14.6141}                  &  0.2226    \\ \cline{2-10} 
                                                      & \multirow{3}{*}{Refinement}                    & (\textit{d}) Base                                         &       0.8615                        &          0.7449                      & \multicolumn{1}{c|}{\textbf{0.8082}}        &     607.8                    &    0.1708                      &     13.7549                      &  0.2082    \\
                                                      &                                                & (\textit{e}) LCL                                          &      0.7506                         &          0.7423                     & \multicolumn{1}{c|}{0.8135}        &          \textbf{604.5}                &          \textbf{0.1283}                &           \textbf{15.0776}                &  \textbf{0.2891}  \\
                                                      &                                                & (\textit{f}) IRP                                          &     \textbf{0.7451}                          &           \textbf{0.7222}                     & \multicolumn{1}{c|}{0.8097}        &         606.4                &          0.1367                &           14.7531                &  0.2350    \\ \hline\hline%\midrule\midrule
    \multirow{6}{*}{DMLab}                            & \multirow{3}{*}{Base}                          & (\textit{g}) Base                                         &         0.9846                      &              1.1798                  & \multicolumn{1}{c|}{1.0803}        &        453.5             &         0.0656            &      17.8548                    &        0.6378      \\
                                                      &                                                & (\textit{h}) DA                                          &      \underline{0.9253}                         &           1.1688                     & \multicolumn{1}{c|}{\underline{0.9818}}        &          \underline{433.5}                &           0.0665               &          17.7900                 &    \underline{0.6408}  \\
                                                      &                                                & (\textit{i}) IRP                                          &       0.9590                        &             \underline{1.0979}                   & \multicolumn{1}{c|}{1.0233}        &        434.5                  &      \underline{0.0642}                    &         \underline{17.9437}                  &   \underline{0.6408}   \\ \cline{2-10} 
                                                      & \multirow{3}{*}{Refinement}                    & (\textit{j}) Base                                         &        0.9946                       &            1.1117                    & \multicolumn{1}{c|}{1.0566}        &      436.4                    &       0.0653                   &      17.8737                     &  0.6342    \\
                                                      &                                                & (\textit{k}) DA                                          &       \textbf{0.9226}                        &                1.1229                & \multicolumn{1}{c|}{\textbf{0.9594}}        &         441.2                 &      0.0668                    &          17.7730                 &    0.6366  \\
                                                      &                                                & (\textit{l}) IRP                                          &          0.9450                     &              \textbf{1.0813}                  & \multicolumn{1}{c|}{1.0045}        &          \textbf{422.7}                &         \textbf{0.0631}                 &             \textbf{18.0227}              &   \textbf{0.6479}   \\ \hline%\bottomrule
    
    \end{tabular}
    }
    \vspace{-4mm}
    \label{tab:main}
    \end{table*}
}

\definecolor{iccvblue}{rgb}{0.21,0.49,0.74}
\usepackage[pagebackref,breaklinks,colorlinks,allcolors=iccvblue]{hyperref}

\def\paperID{2312} % *** Enter the Paper ID here
\def\confName{ICCV}
\def\confYear{2025}

\title{Toward Stable World Models: \\ Measuring and Addressing World Instability in Generative Environments}

\author{Soonwoo Kwon\thanks{Co-first Author $\dagger$ Corresponding Author}
\quad
Jin-Young Kim$^{*}$
\quad 
Hyojun Go$^{1}$
\quad
Kyungjune Baek$^{2\dagger}$
\\ EverEx$^1$ Sejong University$^2$
\\ \tt\small \{swkwon.john, seago0828, gohyojun15\}@gmail.com, kyungjune.baek@sejong.ac.kr}
\begin{document}

\maketitle

\begin{abstract}
We present a novel study on enhancing the capability of preserving the content in world models, focusing on a property we term \textbf{World Stability}. Recent diffusion-based generative models have advanced the synthesis of immersive and realistic environments that are pivotal for applications such as reinforcement learning and interactive game engines. 
However, while these models excel in quality and diversity, they often neglect the preservation of previously generated scenes over time--a shortfall that can introduce noise into agent learning and compromise performance in safety-critical settings. In this work, we introduce an evaluation framework that measures world stability by having world models perform a sequence of actions followed by their inverses to return to their initial viewpoint, thereby quantifying the consistency between the starting and ending observations. Our comprehensive assessment of state-of-the-art diffusion-based world models reveals significant challenges in achieving high world stability. Moreover, we investigate several improvement strategies to enhance world stability.
Our results underscore the importance of world stability in world modeling and provide actionable insights for future research in this domain.

\end{abstract}    
\section{Introduction}
\label{sec:intro}
Recent advancements in generative models, notably diffusion models~\cite{NEURIPS2020_4c5bcfec,Karras2022edm}, have enhanced world models~\cite{world_model_ha2018}, enabling the generation of immersive, realistic environments. 
They serve as engines that generate playable environments~\cite{worldlabs_blog,che2024gamegen} for user or agent, creating simulation setups for reinforcement learning data collection~\cite{hafner2021mastering} and making tasks like robot learning more sample-efficient~\cite{MendoncaRSS23}.

% 밑에 처럼도 써볼 수 있음!! 
% Recent advancements in generative models, such as diffusion models~\cite{NEURIPS2020_4c5bcfec, Karras2022edm}, have significantly improved the quality of world models, enabling the generation of immersive and realistic environments. World models have gained attention for their ability to generate playable environments~\cite{world_model_ha2018} from user or agent inputs, facilitating the creation of realistic simulation environments for reinforcement learning data collection. This advancement enhances sample efficiency in large-scale interaction data collection, reducing the need for labor-intensive data gathering process~\cite{MendoncaRSS23}.

\figteaser
To reliably support these applications with the world model, generated content requires three key characteristics: quality, diversity, and scene preservation. While recent diffusion-based world models~\cite{diamond,ding2024dwm,chen2024diffusion} excel in quality and diversity, previous studies have not paid much attention to the ability to preserve the previously generated environment over time. Despite limited focus, keeping the previously generated scene is a characteristic that significantly differentiates world models from general video generation models. While video generation models aim to synthesize creative videos with generally unidirectional temporal and spatial progression, failing to maintain the previous scene in building an environment can introduce noise into agent learning, leading to slower learning speeds or degradation of the final policy’s quality~\cite{sun2023exploring,huang2024dark,cheng2024rime}. Furthermore, when world models are used to train agents in low fault-tolerance settings such as autonomous driving, lacking this capability may induce incorrect policy and severe safety issues~\cite{zeng2024safety}.

In this work, we introduce the term ``world stability'' to describe the property where, after an agent performs actions and returns to the same location, the environment remains consistent with its initial observation. We inspected existing state-of-the-art (SoTA) methods~\cite{diamond,chen2024diffusion} and found that they struggle significantly with maintaining the environment. 
For example, in the upper image sequence of \Fref{fig:main_figure}, the door on the right disappears upon the agent's return. This can significantly impact the agent's actions, leading to degraded policy learning performance. From a gameplay perspective, it undermines the consistency of the game's progression, severely degrading the user experience. To highlight the problem and provide direction for future researchers, we introduce an evaluation framework and metrics to assess the world stability of various models. Specifically, our framework is applicable when inverse actions are defined within the environment. For example, if an action rotates the agent's view to the left, the inverse action would be rotating it to the right. We design the procedure such that the agent performs the action $N$ times, and then performs the inverse action $N$ times to return to the initial frame. We then compute the consistency between the first frame and the frame generated after the $2N$ actions. Within this framework, we evaluated the state-of-the-art methods’ world stability both quantitatively and qualitatively. 
Furthermore, we explore several potential improvements, including extending the context length, data augmentation by reversing the input sequences, short fine-tuning for reverse modeling, and inference-time scaling using advanced sampling methods. We implement a subset of these approaches and highlight their effectiveness and limitations concerning both generation quality and our newly proposed metric, world stability.

\noindent In summary, our contributions are:
\begin{itemize}
    \item Introducing the novel concept of \textit{world stability} and developing a structured evaluation framework to explicitly measure and emphasize the importance of environment persistency in world modeling.
    \item Evaluating current SoTA diffusion-based world models using our proposed framework and finding the lack of world stability of the SoTA methods.
    \item Investigating several methods to enhance world stability--such as context extension, reversed data augmentation, injection of reverse projection, and inference-time scaling--and analyzing their strengths and limitations.
\end{itemize}
% The proposed method operates in a plug-and-play manner on pre-trained models, making it applicable to existing models. Specifically, it introduces inverse action modeling, where a reverse frame is generated by conditioning on the inverse action, followed by a short fine-tuning process. 
% Then, during the sampling process, for each action taken, the inverse of the previous action is used as a condition to generate a frame. Noise is subsequently added to that frame and it is regenerated to produce a future frame. Through this process, the model is better able to incorporate past information when predicting the future. 
% In addition, we also investigate the way to improve the world stability in the sampling process using more computations in the test time. By doing so, we can trade off the computational complexity and performance.
% We demonstrate that the proposed fine-tuning method and the simple sampling technique indeed improve the world stability of existing pre-trained models through experiments.

\section{Related Works}
\label{sec:rel_work}
\subsection{World Simulation}

World models, originally proposed as simulated environments for training reinforcement learning agents~\cite{world_model_ha2018}, have seen significant evolution. The following works focused on improving the environment's realism to enhance agent policies~\cite{Hafner2020Dream,robine2023transformerbased,zhang2023storm,bruce2024genie}. The concept has since broadened to include interactive virtual environments for user interaction~\cite{Kim_2020_CVPR,valevski2025diffusion}. Recent advances leverage diffusion-based generative models to dramatically improve the quality and diversity of these simulated worlds~\cite{diamond,chen2024diffusion,ding2024dwm,che2024gamegen}. For example, DIAMOND~\cite{diamond} and DWM~\cite{ding2024dwm} utilize action-conditioned diffusion models for improved performance in Atari and locomotion tasks, with DIAMOND capable of learning from datasets without explicit rewards, such as Counter-Strike gameplay footage~\cite{pearce2022counter}. Other approaches, like Diffusion forcing~\cite{chen2024diffusion} that combines diffusion with next-token prediction and GAMEGEN-X~\cite{che2024gamegen} focusing on open-world game engines), demonstrate the versatility of this technology. This work proposes an evaluation protocol for assessing the consistency of these world models and introduces a novel method to improve consistency across SoTA approaches.

\subsection{Evaluation Metrics for World Models}
World models, a specialized type of generative model, possess unique temporal dynamics, action-conditioned generation, and interactivity. Because of its unique characteristics, world models require specialized evaluation metrics. 
A common indirect approach for world models for reinforcement learning is to assess policies trained within the environment~\cite{diamond}, however, this is inapplicable outside RL. As a sidestep, previous works often adopt metrics from generative models, such as Fréchet Inception Distance (FID)~\cite{heusel2017fid} and Fréchet Video Distance (FVD)~\cite{unterthiner2018fvd} which measure distributional distances between real and generated features, LPIPS~\cite{zhang2018lpips} assessing diversity, and CLIP score~\cite{hessel2021clipscore} evaluating action-content alignment.
However, these metrics, designed primarily for images, are insufficient for assessing world models. FVD, despite using video features, is overly influenced by content, neglecting crucial temporal dynamics~\cite{ge2024fvdcontent}. 
Critically, existing metrics fail to assess world stability: the coherent persistence of spatial relationships and object locations over time.
Lack of this consistency, as shown in \cite{zeng2024safety}, can be detrimental in safety-critical domains like autonomous driving.
Recently, MEt3R \cite{asim2025met3rmeasuringmultiviewconsistency} is proposed to evaluate the 3D consistency of generated multi-view images regardless of the quality, content, and camera poses. However, it is not been introduced to the world models yet.
To address limitations of current evaluation that primarily relies on general generative model metrics or indirect policy assessment, we focus on the crucial aspect of world stability, and propose a novel metric for its evaluation.

\section{Proposed Evaluation Framework}
\label{sec:method}
\subsection{Notation}
In this section, we introduce an evaluation protocol for measuring world stability. First, we define the notations to describe the proposed evaluation framework. Since the models are designed for sequential modeling conditioned on past actions and observations (frames), we denote the observation at timestep $i$ as $x_i$ and the corresponding action condition as $a_i$. Similarly, we denote the $i$-th generated frame as $\hat{x}_i$.
For the base models, we can generate a new observation of the next step $\hat{x}_{i+1}$ by feeding $x_i$ and $a_i$ as inputs. 
Additionally, we use $a_i^{-1}$ to represent the inverse action corresponding to $a_i$. For example, if the action $a_n$ denotes ``rotate right'', its inverse action $a_n^{-1}$ corresponds to ``rotate left''. 
%For convenience, sequences of actions and observations within a specific range are denoted using the colon operator ``:'', for example, $x_{1:t}$ represents the set of all observations from step 1 to step $t$. 
We denote the set of all observations from step 1 to $t$ as $x_{1:t}$.
We will use these notations throughout the paper.
\subsection{Evaluation Protocol}
% intro 
Recent advances in generative world models~\cite{diamond, bruce2024genie, valevski2025diffusion, che2024gamegen,Kim_2020_CVPR} have primarily focused on generation quality and controllability, paying less attention to a fundamental aspect to be a world simulator: \textbf{World Stability}. 
As illustrated in~\cref{fig:main_figure}, when an agent executes a sequence of actions and then attempts to return to its initial state, the generated environment often exhibits significant semantic drift. However, no prior work has quantified world stability, limiting improvements in this critical capability.

% Framework
To address this gap, we first introduce the evaluation protocol for quantifying \textbf{World Stability}.
Given an initial state \( x_1 \), we apply a model iteratively over a sequence of $N$ actions, $\mathcal{A} := [a_1, a_2, \dots , a_N]$, transforming the state step by step, producing state sequence $\{\hat{x}_2, \hat{x}_3,...,\hat{x}_{N+1}\}$. 
For reverting the transformation to the initial state, we apply the model again using the corresponding inverse actions, $\mathcal{A}^{-1} := [a_N^{-1}, a_{N-1}^{-1}, \dots , a_1^{-1}]$ in reverse order, yielding the final state $\hat{x}_{2N+1}$. 
For convenience, we reannotate the state sequence $\{x_1, \hat{x}_2,...,\hat{x}_{2N+1}\}$ as $\{x_1, \hat{x}_2,...,\hat{x}_N,\hat{\bar{x}},\hat{x}_N^{\dagger},\hat{x}_{N-1}^\dagger,...,\hat{x}_1^\dagger\}$.
Note that our framework is broadly applicable to any environment and action sequence pair $(\mathcal{A}$, $\mathcal{A}^{-1})$. However, given the increased complexity and stochasticity of mixed action sequences, we restrict experiments to a single action type. For instance, $\mathcal{A}$ may consist of $N$ consecutive $5$-degree left rotations, while $\mathcal{A}^{-1}$ consists of $N$ corresponding $5$-degree right rotations.
By following this framework, we systematically measure the discrepancy between $x_1$ and $\hat{x}_{2N+1}$, which serves as the basis for the quantitative assessment of world stability.

% \figmethod
\figmethodnew
\figsampling

\subsection{World Stability Score}
% \subsection{Evaluation Metric}
% Metic
% Building on the proposed framework, we introduce the \textbf{World Stability (WS) score}, guided by two key principles: (1) \textit{Stability}: The final state after executing an action sequence and its inverse should closely resemble the initial state. This is measured as $\textit{sim}(x_1, \hat{x}_{2N+1})$.  
% (2) \textit{Diversity}: While ensuring consistency, the intermediate states should remain diverse, conditioned on the applied actions. Without this, a model could artificially achieve high consistency by generating nearly identical frames regardless of the action sequence. To prevent this, we assess diversity as $1- \frac{1}{2N} \sum_{i =1}^{2N+1} \left| \textit{sim}(\hat{x}_i, \hat{x}_{N+1}) \right|$, ensuring the model responds appropriately to action conditions.  

Building on the proposed framework, we introduce the \textbf{World Stability (WS) score}, guided by two key principles: (1) \textit{Discrepancy}: After executing an action sequence $\mathcal{A}$ and its inverse $\mathcal{A}^{-1}$, the final state should closely resemble the initial state. We quantify this as $d(x_1, \hat{x}_{1}^\dagger)$, where $d$ represents a distance between two frames. 
% (2) \textit{Diversity}: 
(2) \textit{Dynamics}: 
While ensuring consistency, the intermediate states should reflect the action condition well. Without this constraint, a model could artificially achieve a high stability score by generating nearly identical frames regardless of the action sequence. 
The difficulty of maintaining stability varies depending on the extent of the agent's movement throughout the action sequence—the more it moves, the harder it becomes. To account for this, we define the dynamics as 
% $\frac{1}{2N} \sum_{i =1}^{2N+1} \left| d(\hat{x}_i, \hat{x}_{N+1}) \right|$, 
$\frac{1}{2} (d(x_1, \hat{\bar{x}}) + d(\hat{x}_{1}^\dagger, \hat{\bar{x}}))$, 
% where $\hat{x}_{N/2}$ represents the central frame among the frame sequence.
% ensuring the model appropriately responds to action conditions.  

Finally, we define the WS score as the ratio of discrepancy to dynamics. In formal terms,  
% \begin{equation}
% \label{ws_score}
% \text{WS score} = \frac{d(x_1, \hat{x}_{2N+1})}{\frac{1}{2N}\sum_{i =1}^{2N+1} \left| d(\hat{x}_i, \hat{x}_{N+1}) \right|}

\begin{equation}
\label{ws_score}
\text{WS score} =2 \times \frac{d(x_1, \hat{x}_{1}^\dagger)}{d(x_1, \hat{\bar{x}}) + d(\hat{x}_{1}^\dagger, \hat{\bar{x}})}
\end{equation}

% \begin{equation}
% \label{ws_score}
% \text{WS score} = \frac{DI}{DY},
% \end{equation}
% where
% \begin{equation*}
%  DI = d(x_1, \hat{x}_{2N+1})
% \end{equation*}
% \begin{equation*}
% DY = \frac{1}{2} \left( d(x_1, \hat{x}_{N/2}) + d(\hat{x}_{N+1}, \hat{x}_{N/2}) \right)
% \end{equation*}

% \begin{equation}
% \label{ws_score}
% DI&= d(x_1, \hat{x}_{2N+1}) \\
% DY&= \frac{1}{2} (d(x_1, \hat{x}_{N/2}) + d(\hat{x}_{N+1}, \hat{x}_{N/2})) \\
% \text{WS score} = \frac{\textbf{DI}}{\textbf{DY}}
% \end{equation}
% \begin{equation}
% \label{ws_score}
% \text{WS score} = \frac{\textit{sim}(x_1, \hat{x}_{2N+1})}{1 - \frac{1}{2N}\sum_{i =1}^{2N+1} \left| \textit{sim}(\hat{x}_i, \hat{x}_{N+1}) \right|}
% \end{equation}
This formulation strikes a balance between stability and dynamics. 
A lower WS score is better, as it indicates that the model can reliably return to the initial state after a sequence of consecutive actions while still responding appropriately to each action.
Note that any semantic similarity measure can be used for $d$; in experiments, we employ LPIPS~\cite{zhang2018lpips}, MEt3R~\cite{asim2025met3rmeasuringmultiviewconsistency}, and DINO distance, which we define as $d_{\text{DINO}}(x, y) = 1 - \cos(f(x), f(y))$, 
where $f$ denotes the feature extractor from DINO v2~\cite{oquab2024dinov}, and 
$\cos$ represents cosine similarity. 
Notably, the WS score is reference-free, as it can be computed without requiring a simulator or ground truth frames, making it applicable to a wide range of environments.
In \cref{exp}, we demonstrate that even state-of-the-art world generation models suffer from instability, using the proposed evaluation framework.
% In the following sections, we evaluate state-of-the-art world generation models using the proposed WS score and explore simple methods to improve it. 
% figure
%% seq_len에 따른 metric 변화 observation
% \figmethod

%%%%%%%%%%%%%%%%%%%%%%%%%%%%
%%%%%%%%%%%%%%%%%%%%%%%%%%%%
%%%%%%%%%%%%%%%%%%%%%%%%%%%%
%%%%%%%%%%%%%%%%%%%%%%%%%%%%

% \section{Simple Remedies for World Instability}
\section{Exploring Solutions for World Stability}
\label{method}
% In this section, we discuss several possible approaches to improve world stability of the existing models
In this section, we introduce several potential strategies for improving world stability. We briefly describe each method, highlighting its intuition, expected advantages, and potential limitations. 
Among the feasible solutions, we investigate several approaches, including increasing context length, applying augmentation, incorporating a mechanism to infer reversed observations, and modifying sampling strategies. The quantitative and qualitative evaluations of these strategies are detailed in Section~\ref{exp}.

\subsection{Longer Context Length}
\label{method:LC}
Recent research has explored increasing the context length to generate more consistent and longer videos~\cite{liu2025world,henschel2024streamingt2v}.  
While the consistency problem in video generation differs from the world instability problem, simply extending the context length could help the model retain knowledge about previous states, potentially improving stability.
However, this approach has a major drawback: the computational cost grows exponentially during both training and inference, making it less scalable for long-horizon predictions. 
To better understand this trade-off, we examine how the WS score changes as the context length increases.

% We investigate the change of WS score when the context length is increased.

\subsection{Data Augmentation}
\label{method:Aug}
The most straightforward approach is to construct the training data with sequences where the agent revisits the same states multiple times while performing diverse actions. However, collecting such sequences at scale is challenging. 
To address this, we explore a simple data augmentation technique that leverages existing training data. Specifically, with state and action sequences $\{x_1,...,x_N\}$ and $\{a_1,...,a_N\}$, we append inverse actions $a_i^{-1}$ and their corresponding frames $x_i^\dagger$ to the original sequence, heuristically rewinding the frames in a sequential manner for $N-1$ times from $i=N-1$ to $1$.
% This approach is applied to the DeepMind Lab navigation task~\cite{dmalb}, which consists of only three actions.
For instance, a sequence originally defined as  
$\{(x_1, a_1), (x_2, a_2)\}$ is transformed into $\{(x_1, a_1), (x_2, a_2), (x_3, a_2^{-1}), (x_2, a_1^{-1})\}$.
Despite its benefits, this method is not inherently scalable, as it requires knowledge of the corresponding inverse actions. Defining the inverse of certain actions, such as interacting with the environment (e.g., shooting a gun), might be impossible, and in many environments, multiple actions are executed simultaneously, further complicating the process.

\subsection{Inject Reverse Prediction Capability}
\label{method:Reverse}
We hypothesize that injecting the capability of predicting previous frames under action reversals into a model can enhance world stability.
%, which is pre-trained solely for next-frame generation, 
This objective encourages the model to preserve world knowledge, leading to more coherent long-term dynamics~\cite{chen2017long,li2018video}. 
The challenge in equipping world models with conditional generation on inverse actions lies in the difficulty of obtaining data with exact inverse actions and corresponding frames (e.g., 'shoot' in the CS:GO environment).
% For example, in the CS:GO environment~\cite{pearce2022counter} 'shoot' is a pre-defined action, but the inverse is not explicitly defined, making it challenging to transform the current frame based on an inverse action. Moreover, in many scenarios, multiple inputs are fed into the model simultaneously, further complicating getting exact inverse actions.
To resolve the issue in an action-agnostic manner, we introduce an \textit{inverse action embedding}, which enables the model to process inverse actions alongside standard actions. 
Furthermore, we employ the data augmentation strategy introduced in~\cref{method:Aug} to facilitate learning inverse action conditioning. 
This augmentation strategy ensures that inverse action conditioning can be seamlessly applied to any action type, enhancing the model’s ability to maintain consistency over time.
Although this method requires additional training and a few extra parameters for inverse action embedding, these parameters are not used during inference, ensuring the same inference cost.
%its inference cost is significantly lower than that of the longer-context method introduced in~\cref{method:LC}.
We fine-tune pre-trained world generative models--originally designed for next-frame generation given the original action--with a few additional epochs. Detailed settings and an analysis of the learned inverse action embedding will be provided in~\cref{exp}.

\subsection{Refinement Sampling}
\label{method:Sampling}
In addition to the training methods introduced in~\cref{method:LC,method:Aug,method:Reverse}, we also explore a sampling time refinement technique.
We are inspired by a line of work that utilizes existing images as priors for generation~\cite{meng2022sdedit, saharia2021imagesuperresolutioniterativerefinement}.
These approaches enhance image generation by incorporating information from existing images rather than relying solely on pure noise. By leveraging conditional information, they minimize undesired noise and improve control over the resulting content.
In conventional diffusion-based sampling, the next state is generated as: $x^{t-1} \sim p(x^{t-1} \mid x^t)$, where $x^t$ is the image at the $t$-th denoising step and  $x^0$ is the original image.
As an additional refinement phase, we propose the following steps:

1. Initial Generation:  
   Starting from random noise \( x^T \), iteratively denoise to obtain \( \hat{x}^0 \).

2. Noise Injection:  
   Add Gaussian noise \( \epsilon \) to \( \hat{x}^0 \) to produce a noisy version \( \hat{x}^{0}_{\text{noisy}} \), defined as $\hat{x}^{0}_{\text{noisy}} = \hat{x}_0 + \epsilon.$

3. Refinement:  
   Use \( \hat{x}^{0}_{\text{noisy}} \) as the starting point for another denoising process to obtain the refined image \( \hat{x}^0_{\text{refined}} \).

\noindent This approach enables the model to re-evaluate and refine the generated image, leading to improved quality and stability over temporal sequences. The effect of the proposed refinement sampling is demonstrated in~\cref{fig:qual_csgo,fig:dmlabws}.

\section{Experiments}
\label{exp}
In this section, we evaluate whether state-of-the-art models suffer world instability using our proposed framework and validate the effectiveness of the possible solutions in addressing the instability.
We begin by outlining the experimental setup in \cref{exp_settup}.
Then, in \cref{exp_quantitative}, we present a quantitative evaluation using the proposed evaluation framework, demonstrating that our approach significantly improves stability over the baselines. Additionally, in~\cref{exp:Qualitative},
we validate that the proposed score effectively measures world stability and provides qualitative insights across diverse models.
Finally, in ~\cref{exp_analysis}, we provide detailed analysis for the deeper insight.

\subsection{Experimental Setups}
\label{exp_settup}
In this section, we describe the experimental settings used to evaluate the world instability exhibited by the state-of-the-art world models and the effectiveness of our methods. 
\paragraph{Environments}
Although our evaluation framework is applicable to any environment that features both actions and their corresponding inverse actions, we focus on two complex 3D environments where world instability leads to critical issues: Counter-Strike: Global Offensive (CS:GO)~\cite{pearce2022counter} and DeepMind Lab navigation (DMLab)~\cite{dmalb}.

CS:GO is a popular video game played on the Dust2 map that includes dynamic gameplay and detailed backgrounds with multiple objects. In this context, instability is exemplified by situations where objects present at an initial position vanish after the agent executes a series of actions. Unless otherwise specified, the evaluation frame begins with an action sequence $\mathcal{A}$ consisting of a leftward rotation and is followed by an inverse sequence $\mathcal{A}^{-1}$ involving a rightward rotation, each of which spans 16 actions. For further details on the dataset, please refer to \cite{diamond,pearce2022counter}.

In the DMLab Navigation dataset, which consists of walks in a 3D maze, retracing one's steps is a frequent occurrence. When solving the maze, any change in the map--such as a shift in color or the emergence of unexpected obstacles--can critically impact performance. Detailed information about the dataset can be found in \cite{chen2024diffusion, dmalb}.
% 우리 세팅에 대한 내용 더 넣어야함

% \paragraph{Target Models}
\paragraph{Base Models}
Proposed methods can be applied to diffusion-based world models with minimal change. 
We exploit two pre-trained models within our target environments. 
The first, DIAMOND~\cite{diamond}, utilizes latent space diffusion models and has demonstrated that training on CS:GO can generate playable environments. 
The second, Diffusion Forcing~\cite{chen2024diffusion} balances teacher forcing and autoregressive next-step prediction by assigning distinct noise levels at each timestep. It is trained on the DMLab dataset.
To inject reverse prediction capability, we shortly fine-tune the model with our objective function.
%When training models with injection reverse prediction capability, we apply brief additional training steps with our objective. 
While DIAMOND was originally trained for 800 epochs, we trained only 10 extra epochs, and for DMLab, we reduced 100k steps to 5k.
Our evaluation across both latent and pixel space diffusion models demonstrates the generalizability of our methods.
\tabmain
\paragraph{Evaluation Metrics}
To measure the proposed world stability, we can leverage various metrics to quantify the perceptual similarity between corresponding frames a distance $d$ in \cref{ws_score}.
In our experiments, we employ three metrics to measure the similarity: LPIPS~\cite{zhang2018lpips}, MEt3R~\cite{asim2025met3rmeasuringmultiviewconsistency}, and DINO features~\cite{oquab2024dinov}.
% related work에 dino 넣어줘야하나? no
LPIPS and DINO features are widely recognized metrics for quantifying perceptual similarity between images. % ref?
MEt3R assesses multi-view consistency by warping image content from one view to align with the other. Instead of independently extracting features from each image, MEt3R transforms one image with reference to the other, making it a promising metric for measuring world stability.
Moreover, we utilize commonly used metrics for generation--MSE, PSNR, and SSIM--to measure the discrepancy between $x_1$ and $\hat{x}_{1}^\dagger$ and FVD to assess the quality of the generated frame sequence.

\subsection{Quantitative Results}
\label{exp_quantitative}
% \tabmain
In this section, we apply our evaluation framework to measure world stability in state-of-the-art diffusion-based world models trained in two environments: CS:GO and DMLab. We quantitatively assess their performance and demonstrate the effectiveness of the proposed solutions introduced in~\cref{method}. \cref{tab:main} summarizes the performance of baselines and the proposed approaches. For clarity, we abbreviate each method as \texttt{Sampling Method}--\texttt{Training Method}. We consider two sampling methods—original (\textit{Base}) and refinement (\textit{Refinement})--and four training methods: original (\textit{Base}), longer context length (\textit{LCL}), data augmentation (\textit{DA}), and injecting reverse prediction capability (\textit{IRP}). Note that \textit{DA} is inapplicable to CS:GO, and we found that diffusion forcing with \textit{LCL} is unstable on DMLab.

% our proposed method is effective for both latent and pixel diffusion models across all similarity measures.
% While FVD shows minimal differences, our stability metric highlights significant gaps, demonstrating that FVD fails to capture world stability. % 우리께 맞는 경향이라는걸 보여줘야
% Even without refined sampling, denoted as w/o CRS, fine-tuned model injected reverse prediction capability surpasses the baseline. 

% We also have tried additional baseline based on naive data augmentation strategy. 
% 어떻게 바꿨는지 
% 왜 csgo에는 적용하지 못하였고, general solution이 될 수 없는지
\paragraph{Longer Context Length}
% \tabcontextlength 
% Increasing the context length with additional computation could potentially mitigate world instability.
To investigate the impact of context length on world instability, we train and evaluate a model with an extended context length of 16 (\textit{LCL}) instead of the original 4 (\textit{Base}) on the CS:GO dataset.
\textit{(b) Base-LCL}  achieves a lower WS score than \textit{(a) Base-Base} across all metrics.
However, \textit{(c) Base-IRP}, despite using a shorter context length, outperforms in WS-LPIPS and discrepancy metrics.
Additionally, \textit{(e) Ref-LCL} further improves \textit{LCL} through refinement sampling.
The results suggest that increasing context length is not the only solution to world instability; orthogonal approaches can also be effective.
% Note that we omit the longer context length experiment on DMLab due to the instability of training diffusion forcing with an extended context length.

\paragraph{Data Augmetnation}
We train a model using the augmented training dataset (\cref{method:Aug}). Since inverse actions in CS:GO are nearly impossible to define, we experiment only with three invertible actions on DMLab.
While this data augmentation approach \textit{(h) Base-DA} is not scalable, it significantly improves WS-LPIPS and shows a slight performance gain in other metrics compared to \textit{(g) Base-Base}.

\paragraph{Reverse Modeling}
We train a model to incorporate reverse prediction capability using the proposed data augmentation strategy (\cref{method:Reverse}) across both environments. \textit{(i) Base-IRP} performs slightly worse than explicit data augmentation (\textit{(h) Base-DA}). However, reverse modeling significantly improves performance in CS:GO, where data augmentation is infeasible (\textit{(c) Base-IRP}).
Moreover, \textit{(c) Base-IRP} even outperforms the longer-context model \textit{(b) Base-LCL}, which incurs a substantially higher computational cost during inference. 
These results suggest that enhancing the model with reverse prediction ability is strongly linked to improving world stability.

\paragraph{Refinement Sampling}
Although refinement sampling (\cref{method:Sampling}) doubles the inference time, it consistently improves world stability across all methods, as shown in the rows of \textit{Refinement} (\textit{(d)-(f), (j)-(l)}) compared to \textit{Base} (\textit{(a)-(c), (g)-(i)}). 
Its effectiveness highlights the importance of sampling in generating a world-stable environment. 
A detailed analysis of refinement sampling is provided in the qualitative results ~\cref{exp:Qualitative} and ablation study in~\cref{fig:abl_seqlen}.

\figexample

\subsection{Qualitative Results}
\label{exp:Qualitative}
We present qualitative results to examine the relationship between generated samples and their WS score, as well as to illustrate the improvements with the proposed methods.

As shown in~\cref{fig:qual_csgo}, in the CS:GO settings, when performing a leftward rotation followed by a rightward rotation of equal magnitude at the same position with a sequence length of 16, the baseline model struggles to generate a stable world.
The frames generated by \textit{Base-Base} exhibit unintended shifts, with the door disappearing entirely. In contrast, our proposed methods significantly improve world stability.
\textit{Base-LCL} preserves the door's presence but introduces noticeable distortions. 
\textit{Base-IRP} produces more stable results, and refinement sampling \textit{Ref-IRP} further enhances the alignment with the ground truth, ensuring the door remains in place with improved texture details. 
These qualitative observations strongly align with the WS-LPIPS scores, confirming that the WS-LPIPS score effectively captures world stability.

A similar trend is observed in the DMLab setting, as shown in~\cref{fig:dmlabws}.
% Similar to the above result, in the case of DMLab, qualitative results clearly illustrate significant differences between the baseline model and other approaches, as shown in~\cref{fig:dmlabws}. 
When performing a rotation to the left and subsequently returning to the original viewpoint by rotating right, the baseline model demonstrates visual instability: the picture frame on the wall completely disappears, and an unintended alteration in the color of the floor occurs.
In contrast, other methods effectively address these issues, preserving both structural and color consistency throughout similar rotational movements. 
These qualitative findings closely mirror the WS-DINO scores, reinforcing that this metric effectively reflects world stability.
% These qualitative observations strongly align with the WS-DINO scores, confirming that the WS-DINO score effectively captures world stability.
% This improvement demonstrates the enhanced world stability, contributing directly to a more reliable representation of the environment.

These results collectively demonstrate two key findings: (1) the WS score is a meaningful metric that accurately measures world stability, and (2) the proposed methods can improve world stability.
Due to space constraints, additional qualitative examples are provided in the Appendix.

\figseqlen
\figdmlabws

\subsection{Analysis}
To elucidate the effectiveness of our approach, we present a series of analyses. All studies are conducted on the CS:GO dataset, using pre-trained DIAMOND as a baseline.

\paragraph{Ablation on Sequence Length}
\label{exp_analysis}
The world stability metric consists of two components: discrepancy and dynamics. We analyze how these values change as the action sequence length increases and how this, in turn, affects world stability. To investigate, we conduct experiments by varying the sequence length on CS:GO. Here, the sequence length refers to the length of the action sequence, $|\mathcal{A}|$.

The discrepancy metric naturally degrades as sequence length increases, due to accumulated generation errors and the increased difficulty of the task.
However, both of the training methods, \textit{Base-LCL} and \textit{Base-IRP}, consistently reduce discrepancy while maintaining dynamics similar to the base model, \textit{Base-Base}. 
\textit{Base-LCL} slightly outperforms \textit{Base-IRP} up to its context length (16), corresponding to a sequence length of 8 as the context is twice the sequence length.
These results highlight the strength of \textit{IRP}, as it can retain information about previous states even when they are not explicitly provided as conditions.

\cref{fig:abl_seqlen} describes the effect of \textit{refinement} sampling.
% We investigate the effect of \textit{refinement} sampling in the second column. 
While it slightly degrades the dynamics, it significantly improves discrepancy across sequence lengths and training methods (\textit{Base}, \textit{LCL}, and \textit{IRP}).
The impact of refinement sampling becomes more pronounced as sequence length increases. This effect may be related to its ability to correct object positions and refine details by reducing blurriness, as shown in~\cref{exp:Qualitative}.
The best-performing combination is \textit{Ref-LCL} for short sequence lengths (under 12), while for longer sequences, \textit{Ref-IRP} achieves the lowest WS score.

% As shown in the first row of \cref{fig:abl_seqlen}, the baseline experiences significant degradation beyond a certain point, whereas our method maintains stability more robustly. 
% Due to space constraints, qualitative examples illustrating variations with sequence length are provided in the appendix.
% Similarly, the diversity metric also declines with increasing sequence length. Interestingly, despite conditioning on the inverse of the last action, our method enhances diversity. This is likely because it prevents the model from generating overly blurry or collapsed outputs.
% Finally, in terms of world stability, defined as the ratio of stability to diversity, our approach outperforms the baseline across all metrics and sequence lengths.

\paragraph{Relationship between Learned Action and Inverse Action Embeddings}

% 실험 세팅 이야기 추가 
\textit{IRP} method, described in~\cref{method:Reverse}, trains the model conditioned generation based on the inverse of the action with data augmentation. To analyze the relationship between the original action embeddings and the introduced inverse action embeddings, we visualize the cosine similarity between the learned inverse embeddings and the original embeddings as a heatmap, as shown in \cref{fig:act_analysis}.

In the heatmap, the y-axis represents the actual actions, while the x-axis represents their corresponding inverse actions. For clarity, we focus only on valid action-inverse pairs among diverse actions. As observed in \cref{fig:act_analysis}, diagonal elements—representing each action and its corresponding inverse—consistently exhibit higher similarity.
These findings indicate that our model successfully captures and leverages inverse action relationships through the proposed augmentation and training strategy.
\figactanalysis
% \figablsampling
% \paragraph{Ablation on Sampling Strategy}
% \tabablation
% To evaluate the effectiveness of refinement sampling method described in~\cref{method:Sampling}, we conducted an ablation study on different sampling strategies.

% Increasing the number of sampling steps twice with the baseline (b1) has little effect on the performance.
% When applying our two-stage sampling process without reverse conditioning (b2), we observed some improvement in the WC-MEt3R metric. While this approach performed better than using reverse modeling alone(c1), it still fell short compared to our proposed method(c2).
% These results further highlight the importance of our two-step approach and the effectiveness of the proposed sampling strategy.

% different action pairs

% separated reverse model?

% inverse 로 recon된 이전 프레임 이미지와의 유사도 (appendix?)

\section{Conclusion}
In this work, we introduce world stability, a key yet overlooked aspect of world models, and propose a metric and framework to measure it. We evaluate the diffusion-based world models, showing their struggles with world stability. Several strategies to enhance stability--such as increasing context length, data augmentation, reverse modeling, and improved sampling--are explored and evaluated under the proposed framework. This work positions world stability as a vital evaluation criterion, expected to guide future research, particularly in world models. Challenges remain, including better ways to improve stability across diverse actions, and its impact on agent learning. In summary, this work emphasizes the importance of world stability and lays the foundation for advancements by proposing enhancement methods.

% \subsection{Memory-augmented Modeling} 
% Inspired by memory-augmented deep models~\cite{ravi2024sam2,santoro2016meta,he2024ma}, we explore leveraging explicit memory structures to better preserve previously observed scenes and potentially enhance world stability. However, practical challenges include efficient real-time memory management and accurate recognition of revisited states, known as loop closure detection~\cite{tsintotas2022revisiting}.

{
    \small
    \bibliographystyle{ieeenat_fullname}
    \bibliography{main}
}

\clearpage
\appendix
\onecolumn

\section{Additional Qualitative Results on CS:GO}

\begin{figure*}[h]
    \centering
    \setlength{\tabcolsep}{1pt}
        \begin{tabular}{c c c c c c c}
        %%%%
        \adjincludegraphics[clip,width=0.13\linewidth,trim={0 0 0 0}]{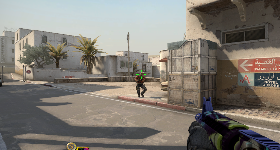} &
        \adjincludegraphics[clip,width=0.13\linewidth,trim={0 0 0 0}]{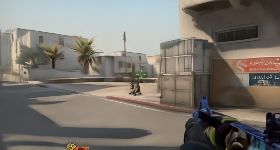} &
        \adjincludegraphics[clip,width=0.13\linewidth,trim={0 0 0 0}]{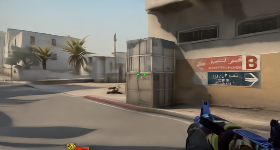} &
        \adjincludegraphics[clip,width=0.13\linewidth,trim={0 0 0 0}]{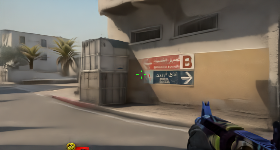} &
        \adjincludegraphics[clip,width=0.13\linewidth,trim={0 0 0 0}]{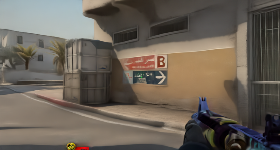}  &

        \adjincludegraphics[clip,width=0.13\linewidth,trim={0 0 0 0}]{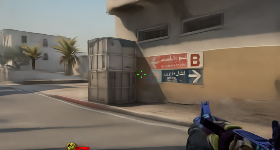}  &
        \adjincludegraphics[clip,width=0.13\linewidth,trim={0 0 0 0}]{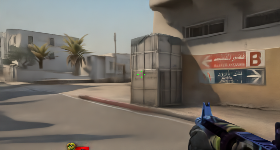} \\ 
        %%%%
        \adjincludegraphics[clip,width=0.13\linewidth,trim={0 0 0 0}]{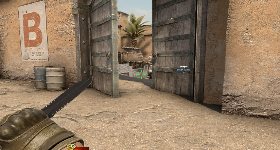} &
        \adjincludegraphics[clip,width=0.13\linewidth,trim={0 0 0 0}]{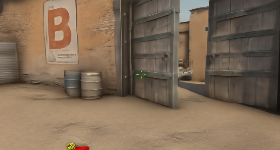} &
        \adjincludegraphics[clip,width=0.13\linewidth,trim={0 0 0 0}]{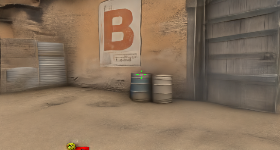} &
        \adjincludegraphics[clip,width=0.13\linewidth,trim={0 0 0 0}]{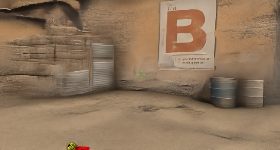} &
        \adjincludegraphics[clip,width=0.13\linewidth,trim={0 0 0 0}]{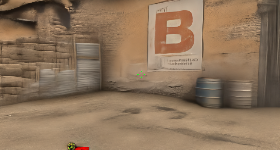}  &
        \adjincludegraphics[clip,width=0.13\linewidth,trim={0 0 0 0}]{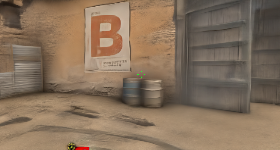}  &
        \adjincludegraphics[clip,width=0.13\linewidth,trim={0 0 0 0}]{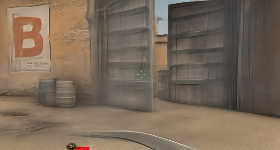} \\ 

        %%%%
        \adjincludegraphics[clip,width=0.13\linewidth,trim={0 0 0 0}]{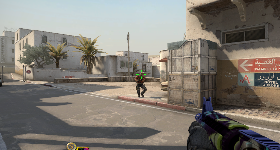} &
        \adjincludegraphics[clip,width=0.13\linewidth,trim={0 0 0 0}]{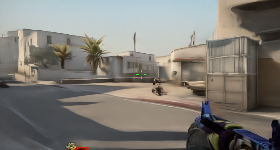} &
        \adjincludegraphics[clip,width=0.13\linewidth,trim={0 0 0 0}]{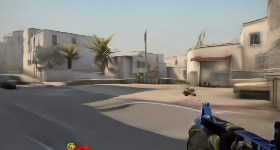} &
        \adjincludegraphics[clip,width=0.13\linewidth,trim={0 0 0 0}]{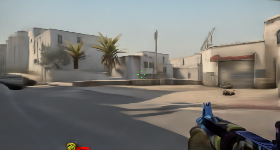} &
        \adjincludegraphics[clip,width=0.13\linewidth,trim={0 0 0 0}]{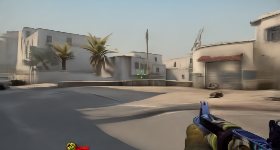}  &
        \adjincludegraphics[clip,width=0.13\linewidth,trim={0 0 0 0}]{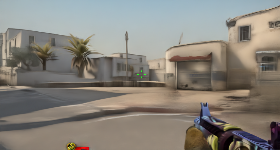}  &
        \adjincludegraphics[clip,width=0.13\linewidth,trim={0 0 0 0}]{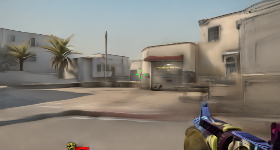} \\ 
        \cmidrule(lr{1pt}){1-7}
        % \specialrule{1.5pt}{0.5pt}{0.5pt}
        \adjincludegraphics[clip,width=0.13\linewidth,trim={0 0 0 0}]{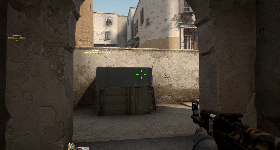} &
        \adjincludegraphics[clip,width=0.13\linewidth,trim={0 0 0 0}]{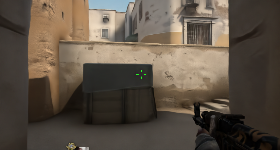} &
        \adjincludegraphics[clip,width=0.13\linewidth,trim={0 0 0 0}]{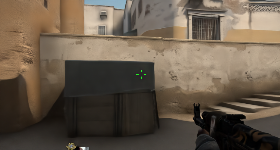} &
        \adjincludegraphics[clip,width=0.13\linewidth,trim={0 0 0 0}]{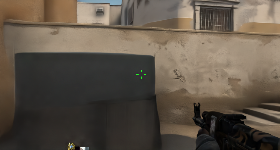} &
        \adjincludegraphics[clip,width=0.13\linewidth,trim={0 0 0 0}]{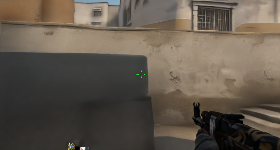}  &

        \adjincludegraphics[clip,width=0.13\linewidth,trim={0 0 0 0}]{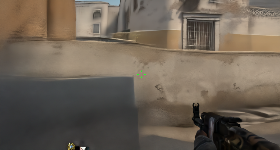}  &
        \adjincludegraphics[clip,width=0.13\linewidth,trim={0 0 0 0}]{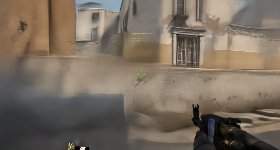} \\ 
        % \rule{\linewidth}{0.5mm} \\
        %%%%
        \adjincludegraphics[clip,width=0.13\linewidth,trim={0 0 0 0}]{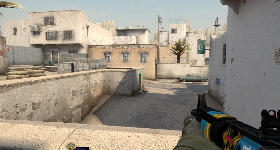} &
        \adjincludegraphics[clip,width=0.13\linewidth,trim={0 0 0 0}]{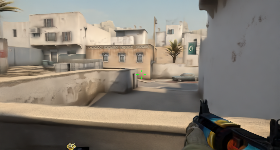} &
        \adjincludegraphics[clip,width=0.13\linewidth,trim={0 0 0 0}]{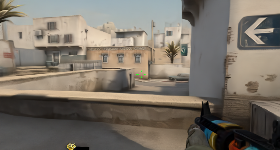} &
        \adjincludegraphics[clip,width=0.13\linewidth,trim={0 0 0 0}]{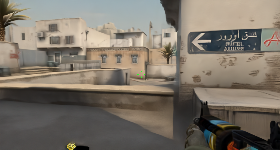} &
        \adjincludegraphics[clip,width=0.13\linewidth,trim={0 0 0 0}]{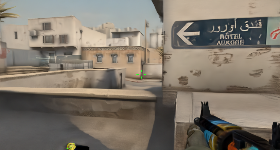}  &
        \adjincludegraphics[clip,width=0.13\linewidth,trim={0 0 0 0}]{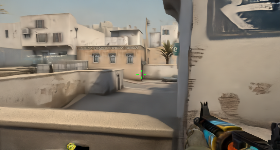}  &
        \adjincludegraphics[clip,width=0.13\linewidth,trim={0 0 0 0}]{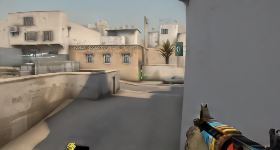} \\ 
        \end{tabular}
    \caption{\textbf{Generated samples following the proposed world stability evaluation framework with action-inverse action pairs.} The first three rows are generated using the left("a" key) and right("b" key) movement. The last two rows correspond to moving forward("w" key) and backward("s" key) in CS:GO game. 
    }
    \vspace{-4mm}
    \label{fig:qual_csgo_appe_acts}
\end{figure*}

We present additional qualitative results on CS:GO. 
First, we illustrate additional generated samples using the proposed evaluation framework. While the main manuscript focuses on the case of rotating the camera left by a certain angle and returning to the original position, here we explore alternative actions in~\cref{fig:qual_csgo_appe_acts}.

\begin{figure*}[h]
    \centering
        % \begin{tabular}{@{\hspace{1mm}}c@{\hspace{1mm}}c@{\hspace{1mm}}c@{\hspace{1mm}}c@{\hspace{1mm}}c@{\hspace{1mm}}c@{\hspace{1mm}}c}
        % \begin{tabular}{@{\hspace{1mm}}c@{\hspace{1mm}} c c c c c c c}
    \setlength{\tabcolsep}{1pt}
        \begin{tabular}{c c c c c c c c}
        %%%%
        \raisebox{0.01\height}{\rotatebox{90}{\shortstack{\scriptsize Base-Base \\ \scriptsize (WS=0.85)}}} &
        \adjincludegraphics[clip,width=0.13\linewidth,trim={0 0 0 0}]{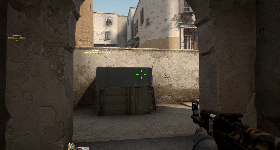} &
        \adjincludegraphics[clip,width=0.13\linewidth,trim={0 0 0 0}]{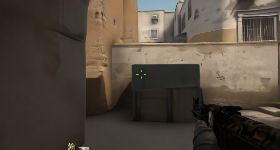} &
        \adjincludegraphics[clip,width=0.13\linewidth,trim={0 0 0 0}]{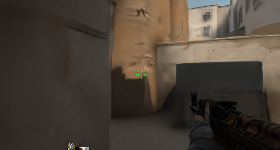} &
        \adjincludegraphics[clip,width=0.13\linewidth,trim={0 0 0 0}]{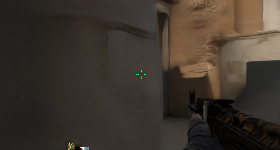} &
        \adjincludegraphics[clip,width=0.13\linewidth,trim={0 0 0 0}]{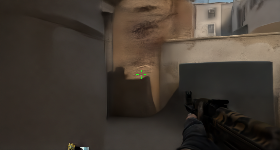}  &

        \adjincludegraphics[clip,width=0.13\linewidth,trim={0 0 0 0}]{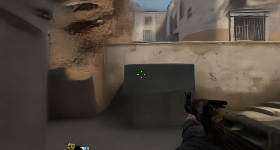}  &
        \adjincludegraphics[clip,width=0.13\linewidth,trim={0 0 0 0}]{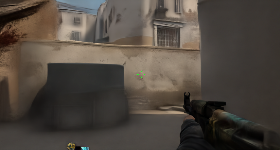} \\ 
        %%%%
        \raisebox{0.01\height}{\rotatebox{90}{\shortstack{\scriptsize Base-LCL \\ \scriptsize (WS=1.08)}}} &
        \adjincludegraphics[clip,width=0.13\linewidth,trim={0 0 0 0}]{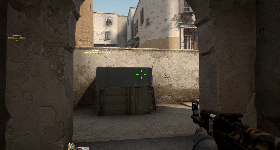} &
        \adjincludegraphics[clip,width=0.13\linewidth,trim={0 0 0 0}]{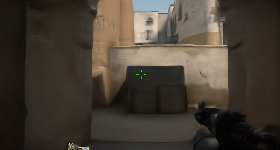} &
        \adjincludegraphics[clip,width=0.13\linewidth,trim={0 0 0 0}]{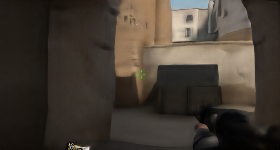} &
        \adjincludegraphics[clip,width=0.13\linewidth,trim={0 0 0 0}]{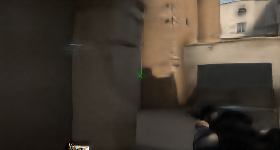} &
        \adjincludegraphics[clip,width=0.13\linewidth,trim={0 0 0 0}]{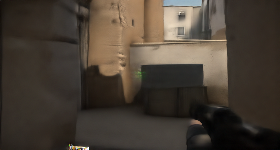}  &

        \adjincludegraphics[clip,width=0.13\linewidth,trim={0 0 0 0}]{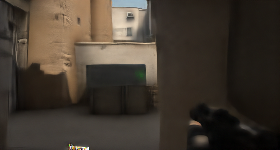}  &
        \adjincludegraphics[clip,width=0.13\linewidth,trim={0 0 0 0}]{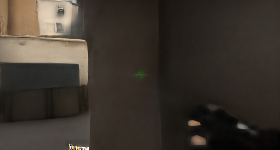} \\ 
        %%%%
        \raisebox{0.01\height}{\rotatebox{90}{\shortstack{\scriptsize Base-IRP \\ \scriptsize (WS=0.98)}}} &
        \adjincludegraphics[clip,width=0.13\linewidth,trim={0 0 0 0}]{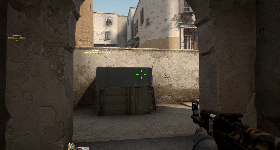} &
        \adjincludegraphics[clip,width=0.13\linewidth,trim={0 0 0 0}]{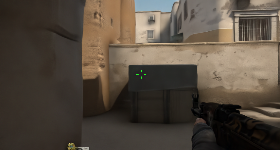} &
        \adjincludegraphics[clip,width=0.13\linewidth,trim={0 0 0 0}]{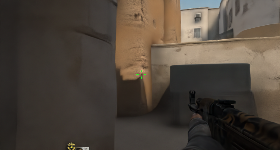} &
        \adjincludegraphics[clip,width=0.13\linewidth,trim={0 0 0 0}]{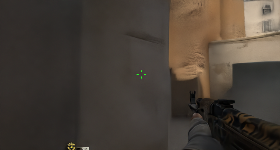} &
        \adjincludegraphics[clip,width=0.13\linewidth,trim={0 0 0 0}]{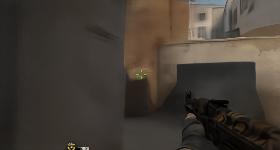}  &

        \adjincludegraphics[clip,width=0.13\linewidth,trim={0 0 0 0}]{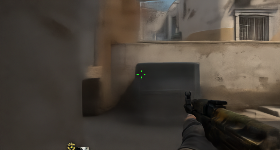}  &
        \adjincludegraphics[clip,width=0.13\linewidth,trim={0 0 0 0}]{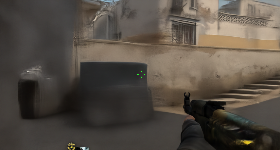} \\
        %%%%
        \raisebox{0.01\height}{\rotatebox{90}{\shortstack{\scriptsize Ref-Base \\ \scriptsize (WS=0.90)}}} &
        \adjincludegraphics[clip,width=0.13\linewidth,trim={0 0 0 0}]{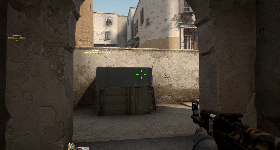} &
        \adjincludegraphics[clip,width=0.13\linewidth,trim={0 0 0 0}]{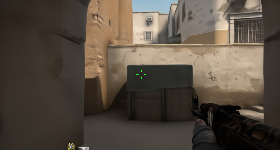} &
        \adjincludegraphics[clip,width=0.13\linewidth,trim={0 0 0 0}]{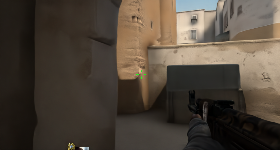} &
        \adjincludegraphics[clip,width=0.13\linewidth,trim={0 0 0 0}]{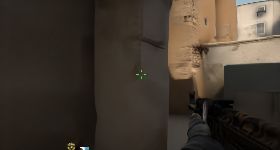} &
        \adjincludegraphics[clip,width=0.13\linewidth,trim={0 0 0 0}]{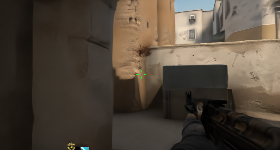}  &

        \adjincludegraphics[clip,width=0.13\linewidth,trim={0 0 0 0}]{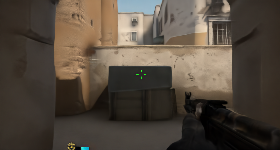}  &
        \adjincludegraphics[clip,width=0.13\linewidth,trim={0 0 0 0}]{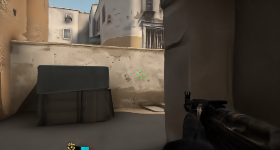} \\
        
        %%%%
        \raisebox{0.01\height}{\rotatebox{90}{\shortstack{\scriptsize Ref-LCL \\ \scriptsize (WS=0.84)}}} &
        \adjincludegraphics[clip,width=0.13\linewidth,trim={0 0 0 0}]{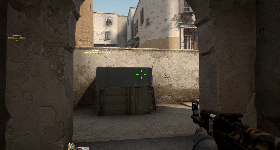} &
        \adjincludegraphics[clip,width=0.13\linewidth,trim={0 0 0 0}]{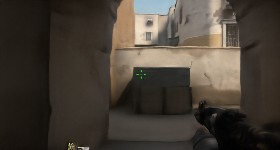} &
        \adjincludegraphics[clip,width=0.13\linewidth,trim={0 0 0 0}]{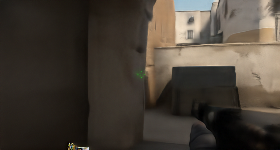} &
        \adjincludegraphics[clip,width=0.13\linewidth,trim={0 0 0 0}]{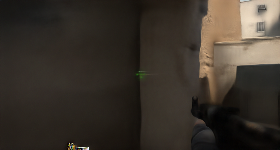} &
        \adjincludegraphics[clip,width=0.13\linewidth,trim={0 0 0 0}]{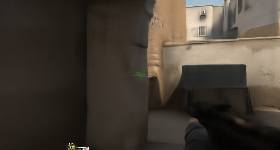}  &

        \adjincludegraphics[clip,width=0.13\linewidth,trim={0 0 0 0}]{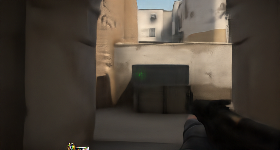}  &
        \adjincludegraphics[clip,width=0.13\linewidth,trim={0 0 0 0}]{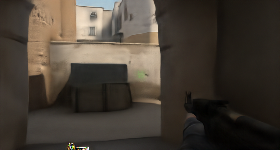} \\
        %%%%
        
        \raisebox{0.01\height}{\rotatebox{90}{\shortstack{\scriptsize Ref-IRP \\ \scriptsize (WS=0.86)}}} &
        \adjincludegraphics[clip,width=0.13\linewidth,trim={0 0 0 0}]{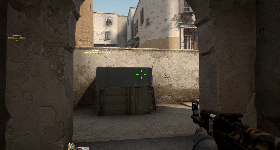} &
        \adjincludegraphics[clip,width=0.13\linewidth,trim={0 0 0 0}]{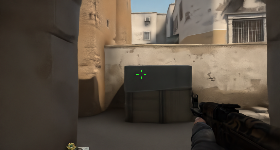} &
        \adjincludegraphics[clip,width=0.13\linewidth,trim={0 0 0 0}]{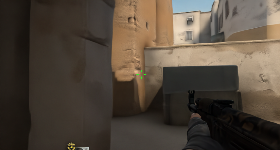} &
        \adjincludegraphics[clip,width=0.13\linewidth,trim={0 0 0 0}]{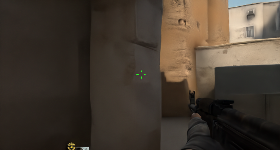} &
        \adjincludegraphics[clip,width=0.13\linewidth,trim={0 0 0 0}]{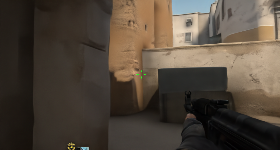}  &

        \adjincludegraphics[clip,width=0.13\linewidth,trim={0 0 0 0}]{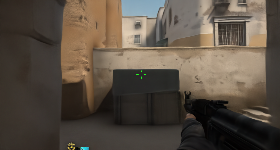}  &
        \adjincludegraphics[clip,width=0.13\linewidth,trim={0 0 0 0}]{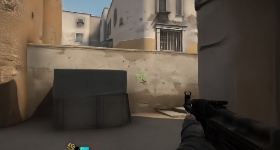} \\

        \end{tabular}
    \caption{\textbf{Qualitative evaluation with the World Stability score on CS:GO.} 
    In the Base model (\textit{Base-Base}), the wall of the building in front becomes noticeably blurry. \textit{Base-LCL} preserves the building but fails to revert to the original state, resulting in a high WS score. While \textit{Base-IRP} correctly captures the viewpoint, the entrance disappears. Refinement sampling (\textit{Ref}) significantly improves all three methods, enhancing both generation quality and world stability.
    % The baseline shows significantly different results compared to the proposed method, while the WS score reflects the ability to preserve previously generated content.
    }
    \vspace{-4mm}
    \label{fig:qual_csgo_appe1}
\end{figure*}

\begin{figure*}[h]
    \centering
        % \begin{tabular}{@{\hspace{1mm}}c@{\hspace{1mm}}c@{\hspace{1mm}}c@{\hspace{1mm}}c@{\hspace{1mm}}c@{\hspace{1mm}}c@{\hspace{1mm}}c}
        % \begin{tabular}{@{\hspace{1mm}}c@{\hspace{1mm}} c c c c c c c}
    \setlength{\tabcolsep}{1pt}
        \begin{tabular}{c c c c c c c c}
        %%%%
        \raisebox{0.01\height}{\rotatebox{90}{\shortstack{\scriptsize Base-Base \\ \scriptsize (WS=0.93)}}} &
        \adjincludegraphics[clip,width=0.13\linewidth,trim={0 0 0 0}]{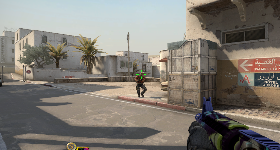} &
        \adjincludegraphics[clip,width=0.13\linewidth,trim={0 0 0 0}]{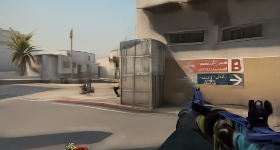} &
        \adjincludegraphics[clip,width=0.13\linewidth,trim={0 0 0 0}]{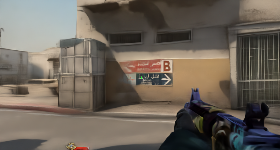} &
        \adjincludegraphics[clip,width=0.13\linewidth,trim={0 0 0 0}]{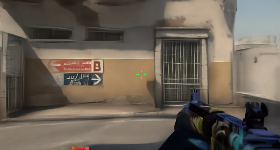} &
        \adjincludegraphics[clip,width=0.13\linewidth,trim={0 0 0 0}]{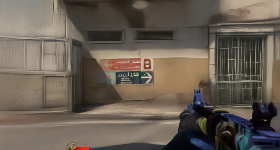}  &

        \adjincludegraphics[clip,width=0.13\linewidth,trim={0 0 0 0}]{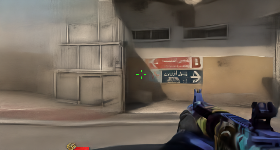}  &
        \adjincludegraphics[clip,width=0.13\linewidth,trim={0 0 0 0}]{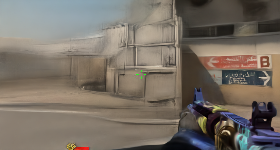} \\ 
        %%%%
        \raisebox{0.01\height}{\rotatebox{90}{\shortstack{\scriptsize Base-LCL \\ \scriptsize (WS=0.67)}}} &
        \adjincludegraphics[clip,width=0.13\linewidth,trim={0 0 0 0}]{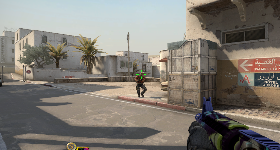} &
        \adjincludegraphics[clip,width=0.13\linewidth,trim={0 0 0 0}]{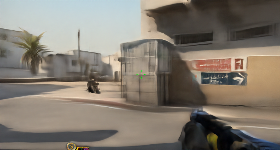} &
        \adjincludegraphics[clip,width=0.13\linewidth,trim={0 0 0 0}]{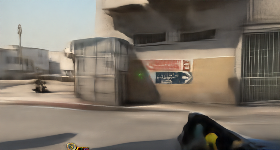} &
        \adjincludegraphics[clip,width=0.13\linewidth,trim={0 0 0 0}]{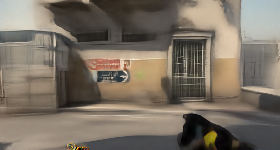} &
        \adjincludegraphics[clip,width=0.13\linewidth,trim={0 0 0 0}]{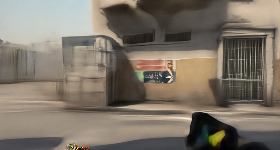}  &

        \adjincludegraphics[clip,width=0.13\linewidth,trim={0 0 0 0}]{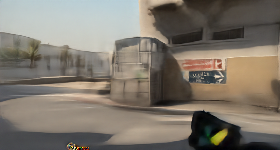}  &
        \adjincludegraphics[clip,width=0.13\linewidth,trim={0 0 0 0}]{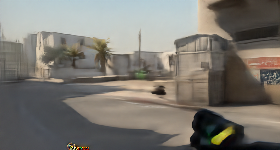} \\ 
        %%%%
        \raisebox{0.01\height}{\rotatebox{90}{\shortstack{\scriptsize Base-IRP \\ \scriptsize (WS=0.82)}}} &
        \adjincludegraphics[clip,width=0.13\linewidth,trim={0 0 0 0}]{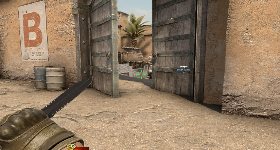} &
        \adjincludegraphics[clip,width=0.13\linewidth,trim={0 0 0 0}]{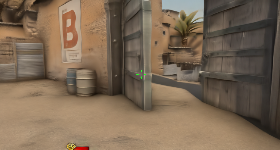} &
        \adjincludegraphics[clip,width=0.13\linewidth,trim={0 0 0 0}]{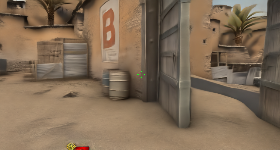} &
        \adjincludegraphics[clip,width=0.13\linewidth,trim={0 0 0 0}]{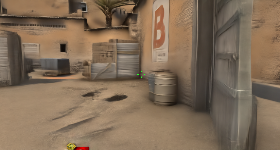} &
        \adjincludegraphics[clip,width=0.13\linewidth,trim={0 0 0 0}]{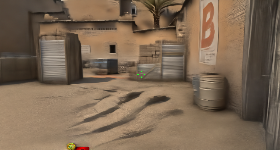}  &

        \adjincludegraphics[clip,width=0.13\linewidth,trim={0 0 0 0}]{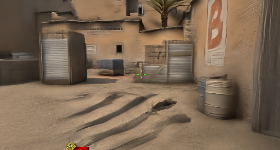}  &
        \adjincludegraphics[clip,width=0.13\linewidth,trim={0 0 0 0}]{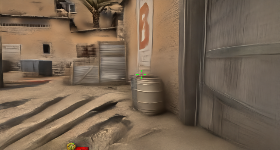} \\
        %%%%
        \raisebox{0.01\height}{\rotatebox{90}{\shortstack{\scriptsize Ref-Base \\ \scriptsize (WS=0.85)}}} &
        \adjincludegraphics[clip,width=0.13\linewidth,trim={0 0 0 0}]{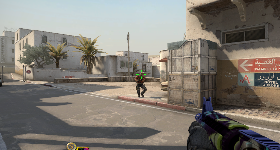} &
        \adjincludegraphics[clip,width=0.13\linewidth,trim={0 0 0 0}]{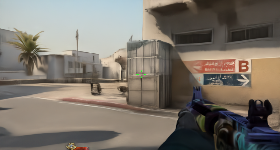} &
        \adjincludegraphics[clip,width=0.13\linewidth,trim={0 0 0 0}]{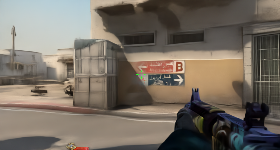} &
        \adjincludegraphics[clip,width=0.13\linewidth,trim={0 0 0 0}]{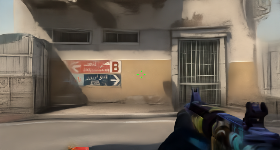} &
        \adjincludegraphics[clip,width=0.13\linewidth,trim={0 0 0 0}]{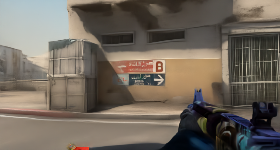}  &

        \adjincludegraphics[clip,width=0.13\linewidth,trim={0 0 0 0}]{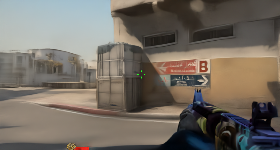}  &
        \adjincludegraphics[clip,width=0.13\linewidth,trim={0 0 0 0}]{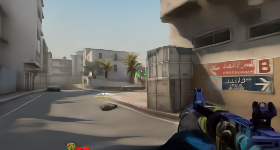} \\
        
        %%%%
        \raisebox{0.01\height}{\rotatebox{90}{\shortstack{\scriptsize Ref-LCL \\ \scriptsize (WS=0.69)}}} &
        \adjincludegraphics[clip,width=0.13\linewidth,trim={0 0 0 0}]{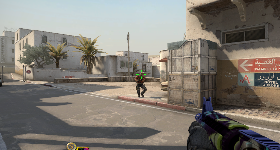} &
        \adjincludegraphics[clip,width=0.13\linewidth,trim={0 0 0 0}]{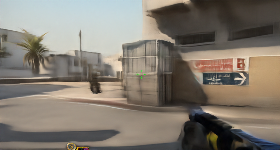} &
        \adjincludegraphics[clip,width=0.13\linewidth,trim={0 0 0 0}]{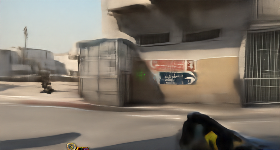} &
        \adjincludegraphics[clip,width=0.13\linewidth,trim={0 0 0 0}]{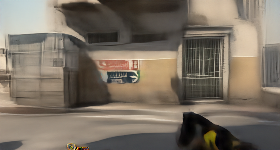} &
        \adjincludegraphics[clip,width=0.13\linewidth,trim={0 0 0 0}]{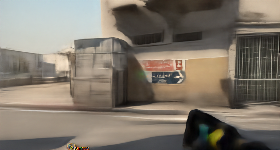}  &

        \adjincludegraphics[clip,width=0.13\linewidth,trim={0 0 0 0}]{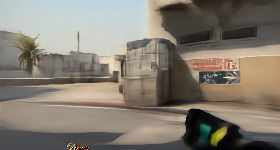}  &
        \adjincludegraphics[clip,width=0.13\linewidth,trim={0 0 0 0}]{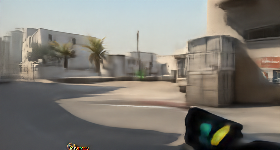} \\
        %%%%
        
        \raisebox{0.01\height}{\rotatebox{90}{\shortstack{\scriptsize Ref-IRP \\ \scriptsize (WS=0.77)}}} &
        \adjincludegraphics[clip,width=0.13\linewidth,trim={0 0 0 0}]{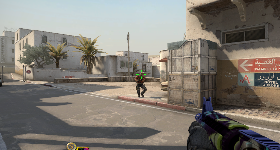} &
        \adjincludegraphics[clip,width=0.13\linewidth,trim={0 0 0 0}]{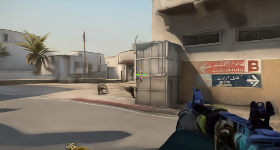} &
        \adjincludegraphics[clip,width=0.13\linewidth,trim={0 0 0 0}]{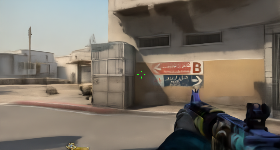} &
        \adjincludegraphics[clip,width=0.13\linewidth,trim={0 0 0 0}]{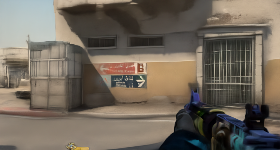} &
        \adjincludegraphics[clip,width=0.13\linewidth,trim={0 0 0 0}]{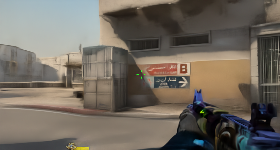}  &

        \adjincludegraphics[clip,width=0.13\linewidth,trim={0 0 0 0}]{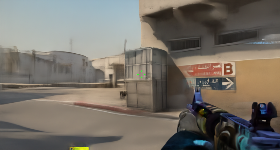}  &
        \adjincludegraphics[clip,width=0.13\linewidth,trim={0 0 0 0}]{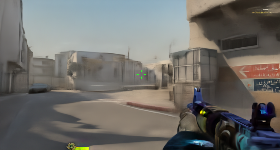} \\

        \end{tabular}
    \caption{\textbf{Qualitative evaluation with the World Stability score on CS:GO.} 
    \textit{Base-Base} fails to return to the original state as the boxes in front of the wll become blurry. 
    While other methods successfully restore the initial viewpoint, the enemy disappears in all cases. Notably, although \textit{Ref-IRP} achieves better generation quality than \textit{Ref-LCL} and \textit{Base-LCL}, the background trees are removed, leading to a worse WS score.
    }
    \vspace{-4mm}
    \label{fig:qual_csgo_appe2}
\end{figure*}

% We also present additional generated examples of the baseline and proposed methods under the same setting as~\cref{fig:qual_csgo}. 
We also present additional generated examples of the baseline and proposed methods under the same setting as Fig. 4. 
These examples not only support our observations described in the manuscript but also include failure cases. The results are shown in~\cref{fig:qual_csgo_appe1,fig:qual_csgo_appe2}.

% Moreover, we provide examples of generated samples with increased sequence length, which were used for experiment results in~\cref{fig:abl_seqlen}.
Moreover, we provide examples of generated samples with increased sequence length, which were used for experimental results in Fig. 5.

\begin{figure*}[h]
    \centering
        % \begin{tabular}{@{\hspace{1mm}}c@{\hspace{1mm}}c@{\hspace{1mm}}c@{\hspace{1mm}}c@{\hspace{1mm}}c@{\hspace{1mm}}c@{\hspace{1mm}}c}
        % \begin{tabular}{@{\hspace{1mm}}c@{\hspace{1mm}} c c c c c c c}
    \setlength{\tabcolsep}{1pt}
        \begin{tabular}{c c c c c c c c}
        %%%%
        \raisebox{0.01\height}{\rotatebox{90}{\shortstack{\scriptsize (SeqLen=8)}}} &
        \adjincludegraphics[clip,width=0.13\linewidth,trim={0 0 0 0}]{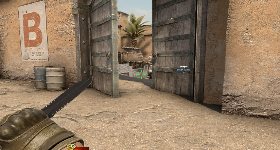} &
        \adjincludegraphics[clip,width=0.13\linewidth,trim={0 0 0 0}]{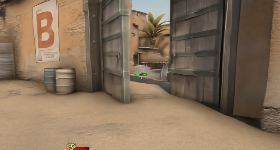} &
        \adjincludegraphics[clip,width=0.13\linewidth,trim={0 0 0 0}]{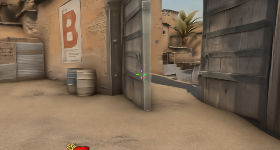} &
        \adjincludegraphics[clip,width=0.13\linewidth,trim={0 0 0 0}]{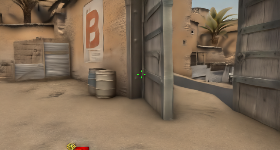} &
        \adjincludegraphics[clip,width=0.13\linewidth,trim={0 0 0 0}]{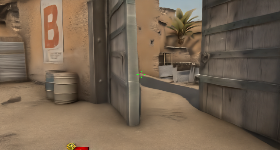}  &

        \adjincludegraphics[clip,width=0.13\linewidth,trim={0 0 0 0}]{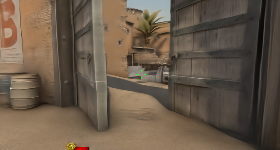}  &
        \adjincludegraphics[clip,width=0.13\linewidth,trim={0 0 0 0}]{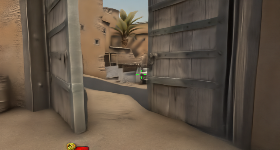} \\ 
        %%%%
        \raisebox{0.01\height}{\rotatebox{90}{\shortstack{\scriptsize (SeqLen=8)}}} &
        \adjincludegraphics[clip,width=0.13\linewidth,trim={0 0 0 0}]{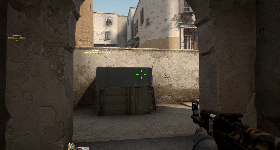} &
        \adjincludegraphics[clip,width=0.13\linewidth,trim={0 0 0 0}]{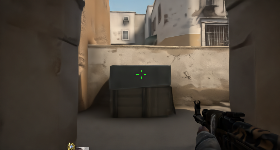} &
        \adjincludegraphics[clip,width=0.13\linewidth,trim={0 0 0 0}]{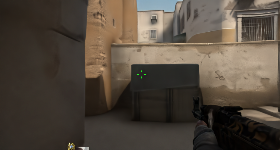} &
        \adjincludegraphics[clip,width=0.13\linewidth,trim={0 0 0 0}]{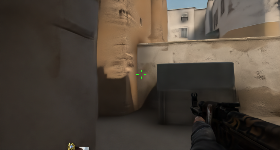} &
        \adjincludegraphics[clip,width=0.13\linewidth,trim={0 0 0 0}]{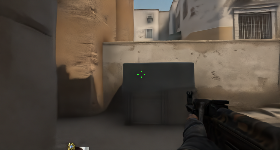}  &

        \adjincludegraphics[clip,width=0.13\linewidth,trim={0 0 0 0}]{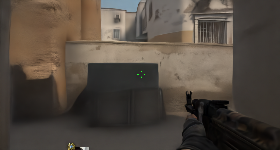}  &
        \adjincludegraphics[clip,width=0.13\linewidth,trim={0 0 0 0}]{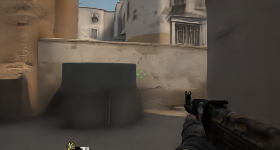} \\ 
        \cmidrule(lr{1pt}){1-7}
        %%%%
        \raisebox{0.01\height}{\rotatebox{90}{\shortstack{\scriptsize (SeqLen=16)}}} &
        \adjincludegraphics[clip,width=0.13\linewidth,trim={0 0 0 0}]{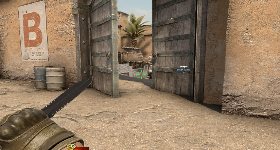} &
        \adjincludegraphics[clip,width=0.13\linewidth,trim={0 0 0 0}]{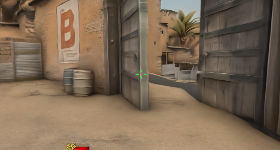} &
        \adjincludegraphics[clip,width=0.13\linewidth,trim={0 0 0 0}]{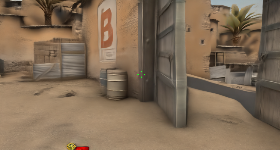} &
        \adjincludegraphics[clip,width=0.13\linewidth,trim={0 0 0 0}]{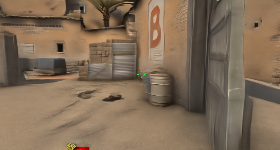} &
        \adjincludegraphics[clip,width=0.13\linewidth,trim={0 0 0 0}]{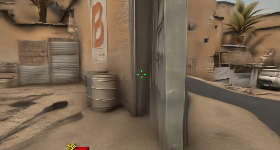}  &

        \adjincludegraphics[clip,width=0.13\linewidth,trim={0 0 0 0}]{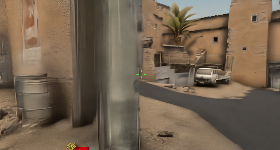}  &
        \adjincludegraphics[clip,width=0.13\linewidth,trim={0 0 0 0}]{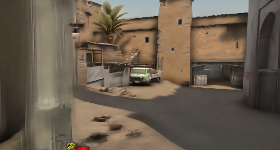} \\ 
        %%%%
        \raisebox{0.01\height}{\rotatebox{90}{\shortstack{\scriptsize (SeqLen=16}}} &
        \adjincludegraphics[clip,width=0.13\linewidth,trim={0 0 0 0}]{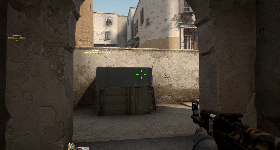} &
        \adjincludegraphics[clip,width=0.13\linewidth,trim={0 0 0 0}]{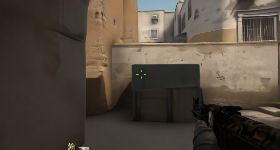} &
        \adjincludegraphics[clip,width=0.13\linewidth,trim={0 0 0 0}]{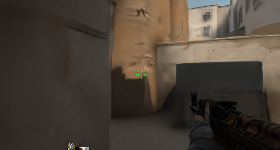} &
        \adjincludegraphics[clip,width=0.13\linewidth,trim={0 0 0 0}]{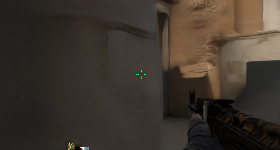} &
        \adjincludegraphics[clip,width=0.13\linewidth,trim={0 0 0 0}]{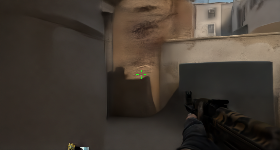}  &

        \adjincludegraphics[clip,width=0.13\linewidth,trim={0 0 0 0}]{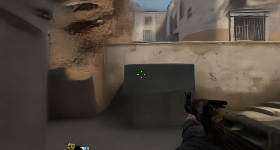}  &
        \adjincludegraphics[clip,width=0.13\linewidth,trim={0 0 0 0}]{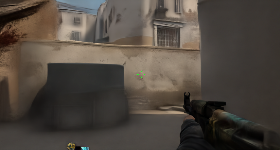} \\ 
        %%%%
        \cmidrule(lr{1pt}){1-7}
        
        \raisebox{0.01\height}{\rotatebox{90}{\shortstack{\scriptsize (SeqLen=24)}}} &
        \adjincludegraphics[clip,width=0.13\linewidth,trim={0 0 0 0}]{figures/qual/csgo_appendix_length/24_1/frame_0.png} &
        \adjincludegraphics[clip,width=0.13\linewidth,trim={0 0 0 0}]{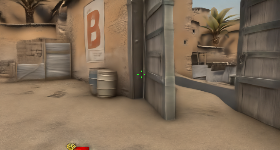} &
        \adjincludegraphics[clip,width=0.13\linewidth,trim={0 0 0 0}]{figures/qual/csgo_appendix_length/24_1/frame_16.png} &
        \adjincludegraphics[clip,width=0.13\linewidth,trim={0 0 0 0}]{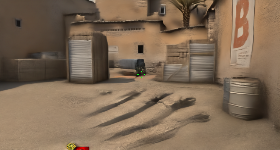} &
        \adjincludegraphics[clip,width=0.13\linewidth,trim={0 0 0 0}]{figures/qual/csgo_appendix_length/24_1/frame_32.png}  &

        \adjincludegraphics[clip,width=0.13\linewidth,trim={0 0 0 0}]{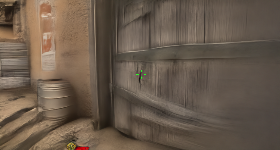}  &
        \adjincludegraphics[clip,width=0.13\linewidth,trim={0 0 0 0}]{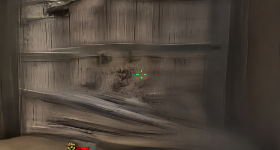} \\ 
        %%%%
        \raisebox{0.01\height}{\rotatebox{90}{\shortstack{\scriptsize (SeqLen=24)}}} &
        \adjincludegraphics[clip,width=0.13\linewidth,trim={0 0 0 0}]{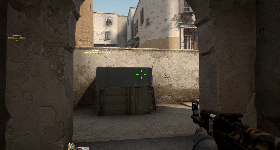} &
        \adjincludegraphics[clip,width=0.13\linewidth,trim={0 0 0 0}]{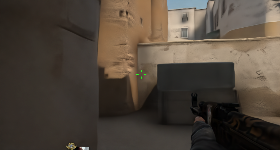} &
        \adjincludegraphics[clip,width=0.13\linewidth,trim={0 0 0 0}]{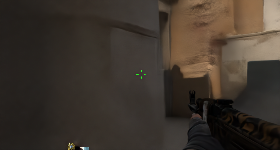} &
        \adjincludegraphics[clip,width=0.13\linewidth,trim={0 0 0 0}]{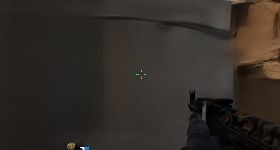} &
        \adjincludegraphics[clip,width=0.13\linewidth,trim={0 0 0 0}]{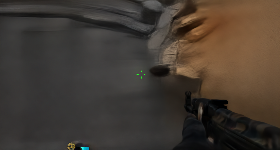}  &

        \adjincludegraphics[clip,width=0.13\linewidth,trim={0 0 0 0}]{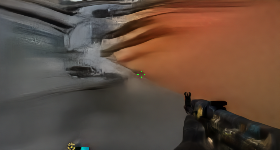}  &
        \adjincludegraphics[clip,width=0.13\linewidth,trim={0 0 0 0}]{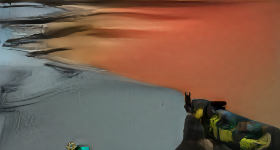} \\ 

        \end{tabular}
    \caption{\textbf{Qualitative evaluation with the World Stability score on CS:GO.} 
    \textit{Base-Base} fails to generate stable world as the sequence length is increased.
    }
    \vspace{-4mm}
    \label{fig:qual_csgo_appe_seq_len}
\end{figure*}

\section{Additional Qualitative Results on DMLab}

\begin{figure*}
    \centering
    \begin{subfigure}[b]{0.33\textwidth}
         \centering
         \includegraphics[width=\textwidth]{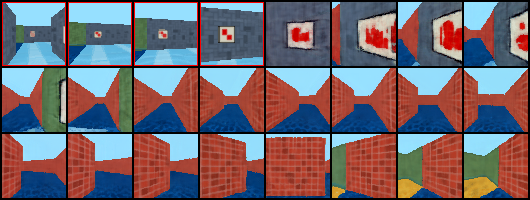}
         \caption{Base-Base}
    \end{subfigure}
    \begin{subfigure}[b]{0.33\textwidth}
         \centering
         \includegraphics[width=\textwidth]{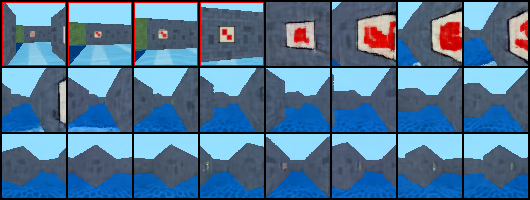}
         \caption{Base-DA}
    \end{subfigure}
    \begin{subfigure}[b]{0.33\textwidth}
         \centering
         \includegraphics[width=\textwidth]{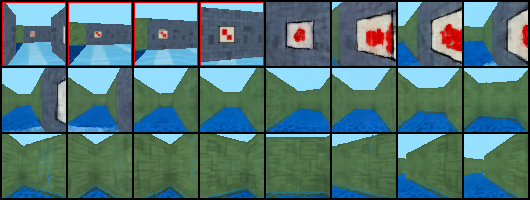}
         \caption{Base-IRP}
    \end{subfigure}
    \begin{subfigure}[b]{0.33\textwidth}
         \centering
         \includegraphics[width=\textwidth]{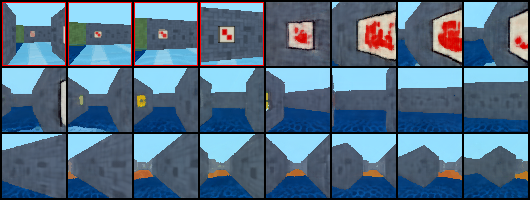}
         \caption{Ref-Base}
    \end{subfigure}
    \begin{subfigure}[b]{0.33\textwidth}
         \centering
         \includegraphics[width=\textwidth]{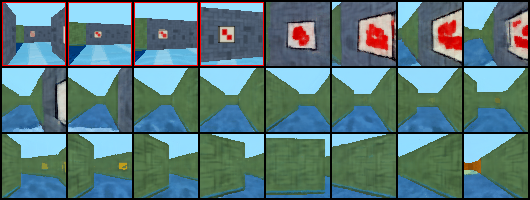}
         \caption{Ref-DA}
    \end{subfigure}
    \begin{subfigure}[b]{0.33\textwidth}
         \centering
         \includegraphics[width=\textwidth]{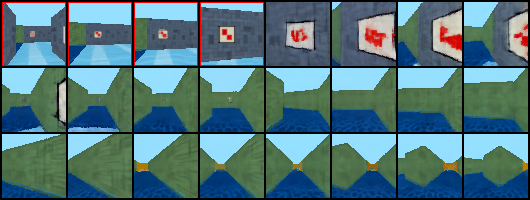}
         \caption{Ref-IRP}
    \end{subfigure}
    \caption{\textbf{Qualitative evaluation with the World Stability score on DMLab.} The action sequence is `forward-left-forward-left'. The Base-Base model violates world stability by generating a red wall that replaced the green wall visible in the context. The Base-DA and Ref-Base models also change the wall color from green to gray. However, other models achieve world stability successfully by maintaining the wall color shown in the context even though the wall is out of sight for a moment.
    } 
    \label{fig:dmlabws_appe1}
\end{figure*}

\begin{figure*}
    \centering
    \begin{subfigure}[b]{0.33\textwidth}
         \centering
         \includegraphics[width=\textwidth]{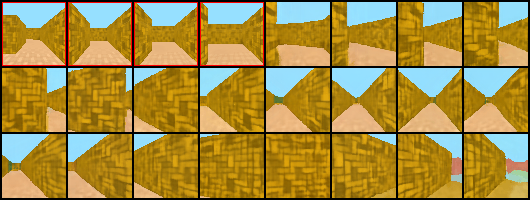}
         \caption{Base-Base}
    \end{subfigure}
    \begin{subfigure}[b]{0.33\textwidth}
         \centering
         \includegraphics[width=\textwidth]{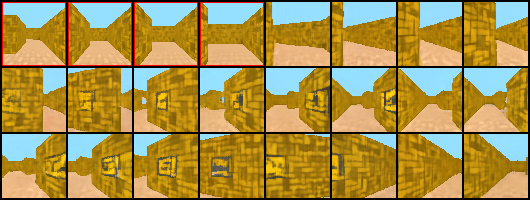}
         \caption{Base-DA}
    \end{subfigure}
    \begin{subfigure}[b]{0.33\textwidth}
         \centering
         \includegraphics[width=\textwidth]{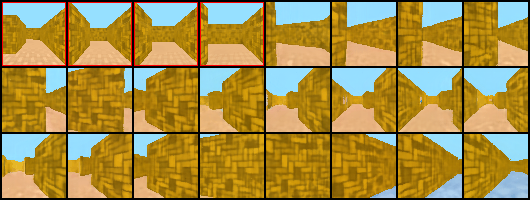}
         \caption{Base-IRP}
    \end{subfigure}
    \begin{subfigure}[b]{0.33\textwidth}
         \centering
         \includegraphics[width=\textwidth]{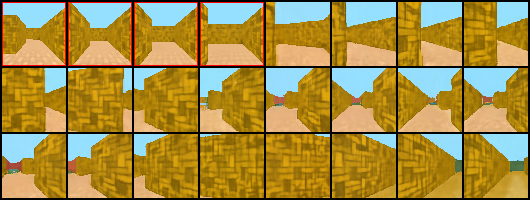}
         \caption{Ref-Base}
    \end{subfigure}
    \begin{subfigure}[b]{0.33\textwidth}
         \centering
         \includegraphics[width=\textwidth]{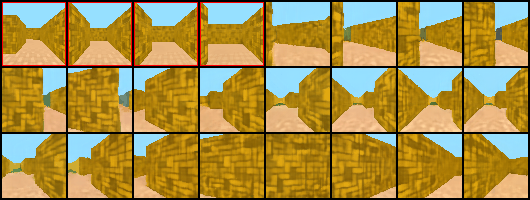}
         \caption{Ref-DA}
    \end{subfigure}
    \begin{subfigure}[b]{0.33\textwidth}
         \centering
         \includegraphics[width=\textwidth]{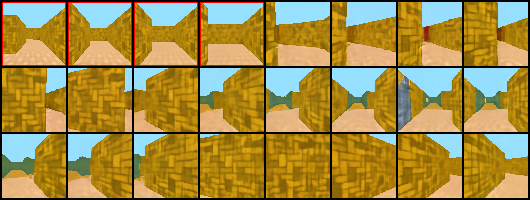}
         \caption{Ref-IRP}
    \end{subfigure}
    \caption{\textbf{Qualitative evaluation with the World Stability score on DMLab.} The action sequence is `forward-left-left-right-right', i.e. `forward-turn around-turn around'. By focusing on the first frame after the context and the last frame, three key points must be stable in this case: wall color, floor color and the road shape. The Base-Base model fails to be stable in all cases. However, after applying the refinement sampling, the floor color can be maintained. Other models except Base-IRP achieve world stability for all three key points.
    } 
    \label{fig:dmlabws_appe2}
\end{figure*}

\begin{figure*}
    \centering
    \begin{subfigure}[b]{0.33\textwidth}
         \centering
         \includegraphics[width=\textwidth]{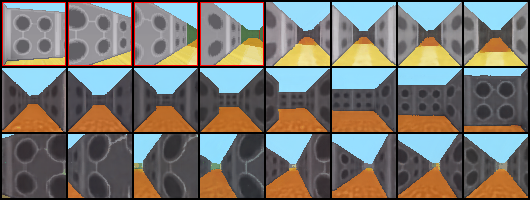}
         \caption{Base-Base}
    \end{subfigure}
    \begin{subfigure}[b]{0.33\textwidth}
         \centering
         \includegraphics[width=\textwidth]{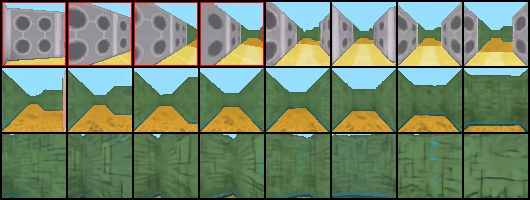}
         \caption{Base-DA}
    \end{subfigure}
    \begin{subfigure}[b]{0.33\textwidth}
         \centering
         \includegraphics[width=\textwidth]{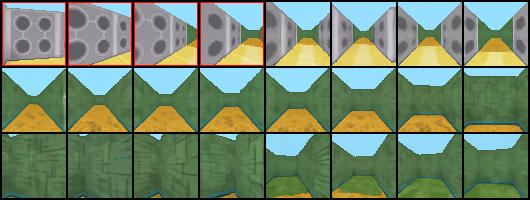}
         \caption{Base-IRP}
    \end{subfigure}
    \begin{subfigure}[b]{0.33\textwidth}
         \centering
         \includegraphics[width=\textwidth]{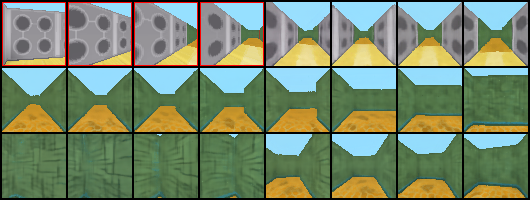}
         \caption{Ref-Base}
    \end{subfigure}
    \begin{subfigure}[b]{0.33\textwidth}
         \centering
         \includegraphics[width=\textwidth]{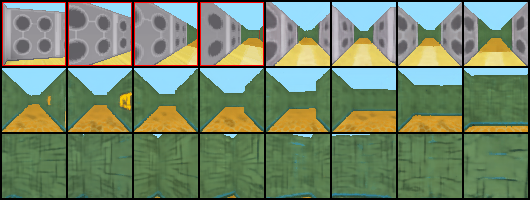}
         \caption{Ref-DA}
    \end{subfigure}
    \begin{subfigure}[b]{0.33\textwidth}
         \centering
         \includegraphics[width=\textwidth]{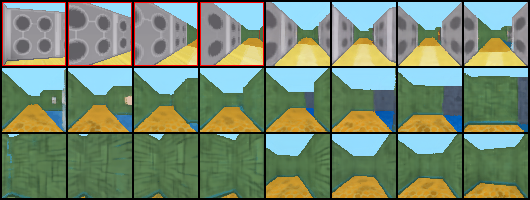}
         \caption{Ref-IRP}
    \end{subfigure}
    \caption{\textbf{Qualitative evaluation with the World Stability score on DMLab.}  The action sequence is `right-forward-forward-left'. This is a case where world stability is lost by hitting the wall while doing `forward' twice. In Base-Base, the color of the wall changes immediately after the context frame.
    } 
    \label{fig:dmlabws_appe3}
\end{figure*}

\begin{figure*}
    \centering
    \begin{subfigure}[b]{0.33\textwidth}
         \centering
         \includegraphics[width=\textwidth]{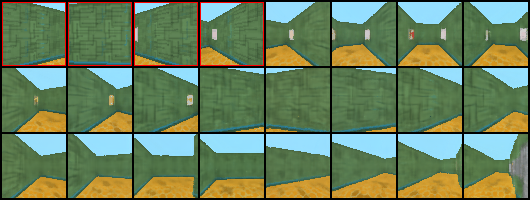}
         \caption{Base-Base}
    \end{subfigure}
    \begin{subfigure}[b]{0.33\textwidth}
         \centering
         \includegraphics[width=\textwidth]{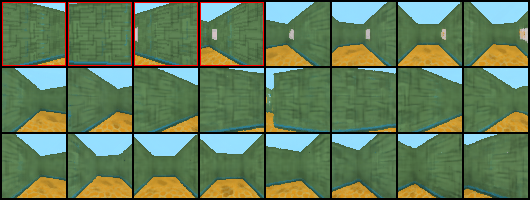}
         \caption{Base-DA}
    \end{subfigure}
    \begin{subfigure}[b]{0.33\textwidth}
         \centering
         \includegraphics[width=\textwidth]{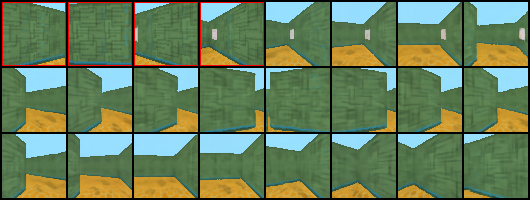}
         \caption{Base-IRP}
    \end{subfigure}
    \begin{subfigure}[b]{0.33\textwidth}
         \centering
         \includegraphics[width=\textwidth]{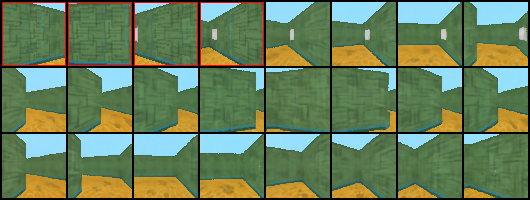}
         \caption{Ref-Base}
    \end{subfigure}
    \begin{subfigure}[b]{0.33\textwidth}
         \centering
         \includegraphics[width=\textwidth]{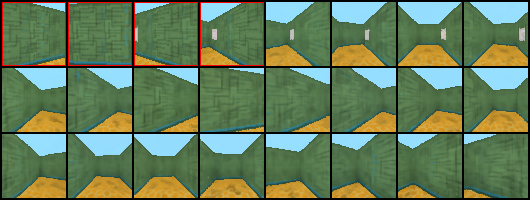}
         \caption{Ref-DA}
    \end{subfigure}
    \begin{subfigure}[b]{0.33\textwidth}
         \centering
         \includegraphics[width=\textwidth]{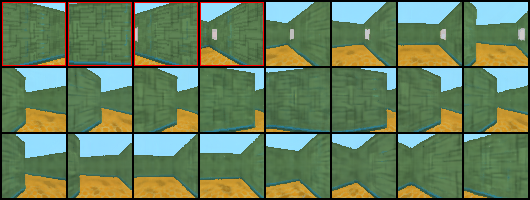}
         \caption{Ref-IRP}
    \end{subfigure}
    \caption{\textbf{Qualitative evaluation with the World Stability score on DMLab.} The action sequence is 'left-left-right-right'. In all cases, the models fail to retain the picture frame on the wall, likely due to its small size. 
    } 
    \label{fig:dmlabws_appe4}
\end{figure*}

We provide additional qualitative results on DMLab in Figures \ref{fig:dmlabws_appe1} and \ref{fig:dmlabws_appe2}. Besides, the failure cases are also illustrated in Figures \ref{fig:dmlabws_appe3} and \ref{fig:dmlabws_appe4}. All examples are generated based on real action sequences present in the test dataset, with the first four frames provided as context.

\section{Additional Possible Apporaches}
In this work, we introduced several ways to improve the world stability including longer context length, data augmentation for generating samples in reverse order, an additional fine-tuning method, and a sampling way to utilize the prior. 
Even if we introduce possible solutions from various perspectives, we introduce the concepts without implementation.

\noindent\textbf{Memory} 
We can utilize memory to store the experiences in a separate field might be one of them. Specifically, inspired by memory-augmented deep models, we can leverage explicit memory structures to preserve previously observed scenes and potentially enhance world stability. However, practical challenges include efficient real-time memory management and accurate recognition of revisited states, known as loop closure detection which is a huge field itself.
% ~\cite{ravi2024sam2,santoro2016meta,he2024ma} ~\cite{tsintotas2022revisiting}

\noindent\textbf{Temporal Coherence Regularization} Temporal coherence regularization adds a penalty to the loss function that discourages abrupt changes or inconsistencies in the model’s outputs or internal representations over time. This can be particularly effective in dynamic environments where maintaining world stability is critical, such as robotics, autonomous navigation, or interactive simulations. The regularization term typically measures the difference between consecutive outputs or states and minimizes this difference unless justified by significant changes in the input data. However, excessive regularization might prevent the model from adapting to legitimate changes in the environment.

% {
%     \small
%     \bibliographystyle{ieeenat_fullname}
%     \bibliography{main}
% }

\end{document}